\documentclass[letterpaper]{article}
\usepackage{uai2020}
\usepackage[margin=1in]{geometry}

\usepackage{microtype}
\usepackage{graphicx}
\usepackage{subfigure}
\usepackage{booktabs} % for professional tables

\usepackage{hyperref}

\usepackage{algpseudocode,algorithm,algorithmicx} 

\usepackage[olditem,oldenum]{paralist}
\usepackage{enumitem}
\usepackage{amsmath}
\usepackage{amsthm}
\usepackage{amssymb}
\usepackage{color}
\usepackage[english]{babel}
\usepackage{graphicx}
\usepackage{grffile}
\usepackage{wrapfig,epsfig}
\usepackage{epstopdf}
\usepackage{url}
\usepackage{color}
\usepackage{epstopdf}
\usepackage{bbm}
\usepackage{comment}
\usepackage{dsfont}


\usepackage[utf8]{inputenc} % allow utf-8 input
\usepackage[T1]{fontenc}    % use 8-bit T1 fonts
\usepackage{booktabs}       % professional-quality tables
\usepackage{amsfonts}       % blackboard math symbols
\usepackage{nicefrac}       % compact symbols for 1/2, etc.
\usepackage{microtype}      % microtypography
\usepackage{multicol}
\usepackage[dvipsnames]{xcolor}
\usepackage[capitalise,nameinlink,noabbrev]{cleveref}

\hypersetup{colorlinks=true,citecolor=blue,linkcolor=blue}

\newcommand{\calS}{\mathcal{S}}
\newcommand{\calA}{\mathcal{A}}

\newcommand{\R}{\mathbb{R}}
\renewcommand{\P}{\mathbb{P}}
\newcommand{\E}{\mathbb{E}}
\newcommand{\KL}{\mathrm{KL}}
\newcommand{\TV}{\mathrm{TV}}

\newcommand{\pr}{\mathrm{Pr}}

\newtheorem{theorem}{Theorem}[section]
\newtheorem{lemma}[theorem]{Lemma}

\usepackage{times}
\usepackage[round]{natbib}
\usepackage{multibib}

\title{Stable Policy Optimization via Off-Policy Divergence Regularization}

\author{Ahmed Touati\textsuperscript{1, $\textbf{*}$}\ \ Amy Zhang \textsuperscript{2, 3, $\textbf{*}$}\  \ Joelle Pineau \textsuperscript{2, 3}\ \  Pascal Vincent\textsuperscript{1, 3} \\
  \textsuperscript{$\textbf{*}$} Equal contribution\ \ \ \ \
  \textsuperscript{1} Mila, Universit\'e de Montr\'eal\ \ \ \ \
  \textsuperscript{2}Mila, McGill University \ \ \ \ \
  \textsuperscript{3}Facebook AI Research
  }

\begin{document}

\maketitle

\begin{abstract}
Trust Region Policy Optimization (TRPO) and Proximal Policy Optimization (PPO) are among the most successful policy gradient approaches in deep reinforcement learning (RL). While these methods achieve state-of-the-art performance across a wide range of challenging tasks, there is room for improvement in the stabilization of the policy learning and how the off-policy data are used. In this paper we revisit the theoretical foundations of these algorithms and propose a new algorithm which stabilizes the policy improvement through a proximity term that constrains the discounted state-action visitation distribution induced by consecutive policies to be close to one another. This proximity term, expressed in terms of the divergence between the visitation distributions, is learned in an off-policy and adversarial manner. We empirically show that our proposed method can have a beneficial effect on stability and improve final performance in benchmark high-dimensional control tasks.
\end{abstract}

\section{INTRODUCTION}
In Reinforcement Learning (RL), an agent interacts with an unknown environment and seeks to learn a policy which maps states to distribution over actions to maximise a long-term numerical reward. Combined with deep neural networks as function approximators, policy gradient methods have enjoyed many empirical successes on RL problems such as video games~\citep{mnih2016asynchronous} and robotics~\citep{levine2016end}. Their recent success can be attributed to their ability to scale gracefully to high dimensional state-action spaces and complex dynamics.

The main idea behind policy gradient methods is to parametrize the policy and perform stochastic gradient ascent on the discounted cumulative reward directly~\citep{sutton2000policy}. To estimate the gradient, we sample trajectories from the distribution induced by the policy. Due to the stochasticity of both policy and environment, variance of the gradient estimation can be very large, and lead to significant policy degradation.  

Instead of directly optimizing the cumulative rewards, which can be challenging due to large variance, some approaches~\citep{kakade2002approximately, azar2012dynamic, pirotta2013safe, schulman2015trust} propose to optimize a surrogate objective that can provide local improvements to the current policy at each iteration. The idea is that the advantage function of a policy $\pi$ can produce a good estimate of the performance of another policy $\pi'$ when the two policies give rise to similar state visitation distributions. Therefore, these approaches explicitly control the state visitation distribution shift between successive policies.

However, controlling the state visitation distribution shift requires measuring it, which is non-trivial. Direct methods are prohibitively expensive. Therefore, in order to make the optimization tractable, the aforementioned methods rely on constraining action probabilities by mixing policies~\citep{kakade2002approximately, pirotta2013safe}, introducing trust regions~\citep{schulman2015trust, achiam2017constrained} or clipping the surrogate objective~\citep{schulman2017proximal, wang2019truly}. 

Our key motivation in this work is that constraining the probabilities of the immediate future actions might not be enough to ensure that the surrogate objective is still a valid estimate of the performance of the next policy and consequently might lead to instability and premature convergence. Instead, we argue that we should reason about the long-term effect of the policies on the distribution of the future states. 

In particular, we directly consider the divergence between state-action visitation distributions induced by successive policies and use it as a regularization term added to the surrogate objective. This regularization term is itself optimized in an adversarial and off-policy manner by leveraging recent advances in off-policy policy evaluation~\citep{nachum2019dualdice} and off-policy imitation learning~\citep{kostrikov2019imitation}. We incorporate these ideas in the PPO algorithm in order to ensure safer policy learning and better reuse of off-policy data. We call our proposed method PPO-DICE.

The present paper is organized as follows: after reviewing conservative approaches for policy learning, we provide theoretical insights motivating our method. We explain how off-policy adversarial formulation can be derived to optimize the regularization term. We then present the algorithmic details of our proposed method. Finally, we show empirical evidences of the benefits of PPO-DICE as well as ablation studies. 

\section{PRELIMINARIES}
\subsection{MARKOV DECISION PROCESSES AND VISITATION DISTRIBUTIONS}
In reinforcement learning, an agent interacts with its environment, which we model as a discounted Markov Decision Process (MDP) $(\calS, \calA, \gamma, \P, r, \rho)$ with state space $\calS$, action space $\calA$, discount factor $\gamma \in [0, 1)$, transition model $\P$ where $\P(s' \mid s, a)$ is the probability of transitioning into state $s'$ upon taking action $a$ in state $s$, reward function $r : (\calS \times \calA) \rightarrow \R$ and initial distribution $\rho$ over $\calS$. We denote by $\pi(a \mid s)$ the probability of choosing action $a$ in state $s$ under the policy $\pi$. The value function for policy $\pi$, denoted $V^{\pi}:\calS \rightarrow \R$, represents the expected sum of discounted rewards along the trajectories induced by the policy in the MDP starting at state $s$: $V^{\pi}(s) \triangleq \E \left [\sum_{t=0}^{\infty} \gamma^t r_t \mid s_0 = s, \pi \right]$. Similarly, the action-value ($Q$-value) function $Q^{\pi}:\calS \times \calA \rightarrow \R$ and the \textit{advantage} function $A^\pi: \calS \times \calA \rightarrow \R$ are defined as: $Q^{\pi}(s, a) \triangleq \E \left [\sum_{t=0}^{\infty} \gamma^t r_t \mid (s_0, a_0)=(s, a), \pi \right]$ and $A^\pi(s, a) \triangleq Q^\pi(s, a) - V^\pi(s)$. 
The goal of the agent is to find a policy $\pi$ that maximizes the expected value from under the initial state distribution $\rho$:
\begin{equation*}
    \max_{\pi} J(\pi) \triangleq (1-\gamma) \E_{s \sim \rho} [V^\pi(s)].
\end{equation*}
We define the discounted state visitation distribution $d^{\pi}_\rho$ induced by a policy $\pi$:
\begin{equation*}
    d^\pi_\rho(s) \triangleq (1 - \gamma) \sum_{t=0}^\infty \gamma^t \pr^\pi(s_t =s \mid s_0 \sim \rho),
\end{equation*}
where $\pr^\pi(s_t =s \mid s_0 \sim \rho)$ is the probability that $s_t = s$, after we execute $\pi$ for $t$ steps, starting from initial state $s_0$ distributed according to $\rho$.
Similarly, we define the discounted state-action visitation distribution $\mu^\pi_\rho(s, a)$ of policy $\pi$
\begin{equation*}
    \mu_\rho^\pi(s, a) \triangleq (1 - \gamma) \sum_{t=0}^\infty \gamma^t \pr^\pi(s_t =s, a_t = a \mid s_0 \sim \rho).
\end{equation*}
It is known \citep{puterman1990markov} that $\mu_\rho^\pi(s, a) = d_\rho^\pi(s) \cdot \pi(a \mid s)$ and that $\mu^\pi$ is characterized via: $\forall (s', a') \in \calS \times \calA$ 
\footnote{By abuse of notation, we confound probability distributions with their Radon–Nikodym derivative with respect to the Lebesgue measure (for continuous spaces) or counting measure (for discrete spaces).}
\begin{align} \label{eq:visitaion_eq}
    \mu^\pi_\rho(s', a') & = (1 -\gamma) \rho(s') \pi(a' \mid s')  \\
    & \quad + \gamma \int \pi(a' \mid s') \P(s' \mid s, a) \mu^\pi_\rho(s, a) ds~da \nonumber,
\end{align}

\subsection{CONSERVATIVE UPDATE APPROACHES}
Most policy training approaches in RL can be understood as updating a current policy $\pi$ to a new improved policy $\pi'$ based on the advantage function $A^\pi$ or an estimate $\hat{A} $ of it. 
We review here some popular approaches that implement conservative updates in order to stabilize policy training.

First, let us state a key lemma from the seminal work of  \citet{kakade2002approximately} that relates the performance difference between two policies to the advantage function. 
\begin{lemma}[The performance difference lemma \citep{kakade2002approximately}] 
\label{lemma:performance diff} 
For all policies $\pi$ and $\pi'$, 
\begin{equation}\label{eq:performance diff}
    J(\pi') = J(\pi) + \E_{s \sim d^{\pi'}_\rho} \E_{a \sim \pi'(. \mid s)} \left[ A^{\pi}(s, a) \right].
\end{equation}
\end{lemma}
This lemma implies that maximizing Equation~\eqref{eq:performance diff} will yield a new policy $\pi'$ with guaranteed performance improvement over a given policy $\pi$. Unfortunately, a naive direct application of this procedure would be prohibitively expensive since it requires estimating $d^{\pi'}_\rho$ for all $\pi'$  candidates.
To address this issue, Conservative Policy Iteration (CPI) \citep{kakade2002approximately} optimizes a surrogate objective defined based on current policy $\pi^i$ at each iteration $i$, 
\begin{equation}
    L_{\pi_i}(\pi') = J(\pi_i) + \E_{\textcolor{red}{s \sim d^{\pi_i}_\rho}} \E_{a \sim \pi'(. \mid s)} \left[ A^{\pi_i}(s, a) \right],
\end{equation}
by ignoring changes in state visitation distribution due to changes in the policy. Then, CPI returns the stochastic mixture $\pi_{i+1} = \alpha_i \pi_i^{+} + (1-\alpha_i) \pi_i$ where $\pi_i^{+} = \arg\max_{\pi'} L_{\pi_i}(\pi')$ is the greedy policy and $\alpha_i \in [0,1]$ is tuned to guarantee a monotonically increasing sequence of policies.

Inspired by CPI, the Trust Region Policy Optimization algorithm (TRPO)~\citep{schulman2015trust} extends the policy improvement step to any general stochastic policy rather than just mixture policies. TRPO maximizes the same surrogate objective as CPI subject to a Kullback-Leibler (KL) divergence constraint
that ensures the next policy $\pi_{i+1}$ stays
within $\delta$-neighborhood of the current policy $\pi_i$:
\begin{align}\label{eq:trpo_update}
    \pi_{i+1} & = \arg\max_{\pi'} L_{\pi_i}(\pi') \\
    \text{s.t} & \quad \E_{s \sim d^{\pi_i}_\rho}\left[ D_{\KL}(\pi'( \cdot \mid s) 
    \| \pi_i( \cdot \mid s))\right] \leq \delta,
    \nonumber
\end{align}
where $D_{\KL}$ is the Kullback–Leibler divergence. In practise, TRPO considers a differentiable parameterized policy $\{ \pi_\theta, \theta \in \Theta\}$ and solves the constrained problem~\eqref{eq:trpo_update} in parameter space $\Theta$. In particular, the step direction
is estimated with conjugate gradients, which requires the computation of multiple Hessian-vector products. Therefore, this step can be computationally heavy.

To address this computational bottleneck, Proximal Policy Optimization (PPO) \citep{schulman2017proximal}
proposes replacing the KL divergence constrained objective~\eqref{eq:trpo_update} of TRPO by
clipping the objective function directly as:
\begin{align}\label{eq:ppo_update}
    L^{\mathrm{clip}}_{\pi_i}& (\pi') = \E_{ (s, a) \sim  \mu^{\pi_i}_\rho}\Big[ \min\big\{ A^{\pi_i}(s, a) \cdot \kappa_{\pi'/\pi_i}(s, a),  \nonumber \\
    & A^{\pi_i}(s, a) \cdot \mathrm{clip}(\kappa_{\pi'/\pi_i}(s, a), 1-\epsilon, 1+\epsilon)\big\}\Big], 
\end{align}
where $\epsilon >0$ and $\kappa_{\pi'/\pi_i}(s, a) = \frac{\pi'(s, a)}{\pi_i(s, a)}$ is the importance 
sampling ratio.

\section{THEORETICAL INSIGHTS}
In this section, we present the theoretical motivation of our proposed method.

At a high level, algorithms CPI, TRPO, and PPO follow similar policy update schemes. They optimize some surrogate performance objective ($L_{\pi_i}(\pi')$ for CPI and TRPO and $L^{\mathrm{clip}}_{\pi}(\pi')$ for PPO) while ensuring that the new policy $\pi_{i+1}$ stays in the vicinity of the current policy $\pi_i$. The vicinity requirement is implemented in different ways:
\begin{compactenum}[\hspace{0pt} 1.]
    \setlength{\itemsep}{2pt}
    \item CPI computes a sequence of stochastic policies that are mixtures between consecutive greedy policies.
    \item TRPO imposes a constraint on the KL divergence between old policy and new one ($ \E_{s \sim d^{\pi_i}_\rho}\left[ D_{\KL}(\pi'( \cdot \mid s) \| \pi_i( \cdot \mid s))\right] \leq \delta$).
    \item PPO directly clips the objective function based on the value of the importance sampling ratio $\kappa_{\pi'/\pi_i}$ between the old policy and new one.
\end{compactenum}
Such conservative updates are critical for the stability of the policy optimization. In fact, the surrogate objective $L_{\pi_i}(\pi')$ (or its clipped version) is valid only in the neighbourhood of the current policy $\pi_i$, i.e, when $\pi'$ and $\pi_i$ visit all the states with similar probabilities. The following lemma more precisely formalizes this\footnote{The result is not novel, it can be found as intermediate step in proof of theorem 1 in \citet{achiam2017constrained}, for example.}: %make this more precise:

\begin{lemma} 
\label{lemma:lower_bound_perf} For all policies $\pi$ and $\pi'$,
\begin{align} \label{eq:lower_bound_perf}
    J(\pi') & \geq L_\pi(\pi') - \epsilon^\pi D_{\TV}(d^{\pi'}_\rho \| d^{\pi}_\rho) \\
    & \geq L^{\mathrm{clip}}_\pi(\pi') - \epsilon^\pi D_{\TV}(d^{\pi'}_\rho \| d^{\pi}_\rho), \nonumber 
\end{align}
where $\epsilon^\pi = \max_{s \in \calS} |\E_{a \sim \pi'(\cdot \mid s)} \left[ A^{\pi}(s, a) \right]|$ and $D_{\TV}$ is the total variation distance.
\end{lemma}
The proof is provided in appendix for completeness.
Lemma~\ref{lemma:lower_bound_perf} states that $L_\pi(\pi')$ (or $L^{\mathrm{clip}}_\pi(\pi')$) is a sensible lower bound to $J(\pi')$ as long as $\pi$ and $\pi'$ are close in terms of total variation distance between their corresponding state visitation distributions $d^{\pi'}_\rho$ and $d^\pi_\rho$. However, the aforementioned approaches enforce closeness of $\pi'$ and $\pi$ in terms of their action probabilities rather than their state visitation distributions. This can be justified by the following inequality~\citep{achiam2017constrained}:
\begin{equation} \label{eq:bound_tv_div}
    D_{\TV}(d^{\pi'}_\rho\| d^{\pi}_\rho) \leq \frac{2 \gamma}{1-\gamma} \E_{s \sim d^\pi_\rho} \left[ D_{\TV}( \pi'(. | s) \| \pi(. | s)) \right].
\end{equation}
Plugging the last inequality~\eqref{eq:bound_tv_div} into~\eqref{eq:lower_bound_perf} leads to the following lower bound:
\begin{equation}\label{eq:bound_pol}
    J(\pi') \geq L_\pi(\pi') - \frac{2\gamma\epsilon^\pi}{1-\gamma} \E_{s \sim d^\pi_\rho} \left[ D_{\TV}( \pi'(. | s) \| \pi(. | s)) \right].
\end{equation}
The obtained lower bound~\eqref{eq:bound_pol} is, however, clearly looser than the one in inequality \eqref{eq:bound_tv_div}. Lower bound~\eqref{eq:bound_pol} suffers from an additional multiplicative factor $\frac{1}{1-\gamma}$, which is the effective planning horizon. It is essentially due to the fact that we are characterizing a long-horizon quantity, such as the state visitation distribution $d^\pi_\rho(s)$, by a one-step quantity, such as the action probabilities $\pi(\cdot \mid s)$. Therefore, algorithms that rely solely on action probabilities to define closeness between policies should be expected to suffer from  instability and premature convergence in long-horizon problems.

Furthermore, in the exact case if we take at iteration $i$, $\pi_{i+1} \leftarrow \arg \max_{\pi'} L_{\pi_i}(\pi') - \epsilon^{\pi_i} D_{\TV}(d^{\pi'}_\rho \| d^{\pi_i}_\rho)$, then
\begin{align*}
    J(\pi_{i+1}) & \geq L_{\pi_i}(\pi_{i+1}) - \epsilon^{\pi_i} D_{\TV}(d^{\pi_{i+1}}_\rho \| d^{\pi_i}_\rho) \\
    & \geq L_{\pi_i}(\pi_{i}) \tag{by optimality of $\pi_{i+1}$}\\
    & = J(\pi_i)
\end{align*}
Therefore, this provides a monotonic policy improvement, while TRPO suffers from a performance degradation that depends on the level of the trust region $\delta$ (see Proposition 1 in~\citet{achiam2017constrained}).

It follows from our discussion that $D_{\TV}(d^{\pi'}_\rho \| d^{\pi}_\rho)$ is a more natural proximity term to ensure safer and more stable policy updates. Previous approaches excluded using this term because we don't have access to $d^{\pi'}_\rho$, which would require executing $\pi'$ in the environment. In the next section, we show how we can leverage recent advances in off-policy policy evaluation to address this issue.

\section{OFF-POLICY FORMULATION OF DIVERGENCES}
In this section, we explain how divergences between state-visitation distributions can be approximated. This is done by leveraging ideas from recent works on off-policy learning ~\citep{nachum2019dualdice, kostrikov2019imitation}.

Consider two different policies $\pi$ and $\pi'$. Suppose that we have access to state-action samples generated by executing the policy $\pi$ in the environment, i.e, $(s, a) \sim \mu_\rho^\pi$. As motivated by the last section, we aim to estimate $D_{\TV}(d^{\pi'}_\rho \| d^{\pi}_\rho)$ without requiring on-policy data from $\pi'$. Note that in order to avoid using importance sampling ratios, it is more convenient to estimate $D_{\TV}(\mu^{\pi'}_\rho \| \mu^{\pi}_\rho)$, i.e, the total divergence between state-action visitation distributions rather than the divergence between state visitation distributions. This is still a reasonable choice as $D_{\TV}(d^{\pi'}_\rho \| d^{\pi}_\rho)$ is upper bounded by $D_{\TV}(\mu^{\pi'}_\rho \| \mu^{\pi}_\rho)$ as shown below:
\begin{align*}
    D_{\TV}(d^{\pi'}_\rho \| d^{\pi}_\rho) & = \int_s \Big|(d^{\pi'}_\rho - d^{\pi}_\rho)(s) \Big| ds \\ 
    & = \int_s \Big | \int_a (\mu^{\pi'}_\rho - \mu^{\pi}_\rho)(s, a) da \Big | ds  \\
& \leq \int_s \int_a \Big| (\mu^{\pi'}_\rho - \mu^{\pi}_\rho)(s, a)\Big| da~ds \\ & =D_{\TV}(\mu^{\pi'}_\rho \| \mu^{\pi}_\rho).
\end{align*}
The total variation distance belongs to a broad class of divergences known as $\phi$-divergences~\citep{sriperumbudur2009integral}. A $\phi$-divergence is defined as,
\begin{equation}\label{eq:phi_div}
    D_\phi(\mu^{\pi'}_\rho \| \mu^{\pi}_\rho) = 
    \E_{(s, a) \sim \mu^{\pi'}_\rho} \left[ \phi\left (\frac{\mu^{\pi}_\rho(s, a)}{\mu^{\pi'}_\rho(s, a)}\right) \right],
\end{equation}
where $\phi: [0, \infty) \rightarrow \R$ is a convex, lower-semicontinuous function and $\phi(1)= 0$. Well-known divergences can be obtained by appropriately choosing $\phi$. These include the KL divergence ($\phi(t) = t \log(t)$), total variation distance ($\phi(t) = |t-1|$), $\chi^2$-divergence ($\phi(t) = (t-1)^2$), etc.
Working with the form of $\phi$-divergence given in Equation~\eqref{eq:phi_div} requires access to analytic expressions of both $\mu^{\pi}_\rho$
and $\mu^{\pi}_\rho$ as well as the ability to sample from $\mu^{\pi'}_\rho$. We have none of these in our problem of interest. To bypass these difficulties, we turn to the alternative variational representation of $\phi$-divergences~\citep{nguyen2009surrogate, huang2017parametric} as
\begin{align}\label{eq:variational_div}
    D_\phi(\mu^{\pi'}_\rho \| \mu^{\pi}_\rho)  = & \sup_{f: \calS \times \calA \rightarrow \R}  \Big[ \E_{(s, a) \sim \mu^{\pi'}_\rho}[ f(s, a)]- \nonumber \\
    & \quad \E_{(s, a) \sim \mu^{\pi}_\rho}[ \phi^\star \circ f(s, a)] \Big], 
\end{align}
where $\phi^\star(t) = \sup_{u \in \R}\{t u - \phi(u)\}$ is the convex conjugate of $\phi$. 
The variational form in Equation~\eqref{eq:variational_div} still requires sampling from $\mu^{\pi'}_\rho$, which we cannot do. To address this issue, we use a clever change of variable trick introduced  by~\citet{nachum2019dualdice}. Define $g:\calS \times \calA \rightarrow \R$ as the fixed point of the following Bellman equation,
\begin{equation} \label{eq:change_var}
    g(s, a) = f(s, a) + \gamma \P^{\pi'}g(s, a),
\end{equation}
where $\P^{\pi'}$ is the transition operator induced by $\pi'$, defined as $\P^{\pi'}g(s, a) = \int \pi'(a' \mid s') \P(s' \mid s, a) g(s', a')$. $g$ may be interpreted as the action-value function of the policy $\pi'$ in a modified MDP which shares the same transition model $\P$ as the original MDP, but has $f$ as the reward function instead of $r$. Applying the change of variable~\eqref{eq:change_var} to \eqref{eq:variational_div} and after some algebraic manipulation as done in \citet{nachum2019dualdice}, we obtain
\begin{align}\label{eq:dice}
    D_\phi(\mu^{\pi'}_\rho \| & \mu^{\pi}_\rho)  = \sup_{g: \calS \times \calA \rightarrow \R}  \Big[ (1-\gamma) \E_{s \sim \rho, a \sim \pi'}[ g(s, a)]- \nonumber \\
    & \E_{(s, a) \sim \mu^{\pi}_\rho}\left[ \phi^\star  \left((g - \gamma \P^{\pi'}g)(s,a)\right)\right] \Big].
\end{align}
Thanks to the change of variable, the first expectation over $\mu^{\pi'}_\rho$ in \eqref{eq:variational_div} is converted to an expectation over the initial distribution and the policy i.e $s \sim \rho(\cdot),  a \sim \pi'(\cdot \mid s)$. Therefore, this new form of the $\phi$-divergence in \eqref{eq:dice} is completely off-policy and can be estimated using only samples from the policy $\pi$.

\paragraph{Other possible divergence representations:} Using the variational representation of $\phi$-divergences was a key step in the derivation of Equation~\eqref{eq:dice}. But in fact any representation that admits a linear term with respect to $\mu^{\pi'}_\rho$ (i.e $\E_{(s, a) \sim \mu^{\pi'}_\rho}[ f(s, a)]$) would work as well. For example, one can use the the Donkser-Varadhan representation~\citep{donsker1983asymptotic} to alternatively express the KL divergence as:
\begin{align}\label{eq:donkser}
    D_\phi(\mu^{\pi'}_\rho \| \mu^{\pi}_\rho)  = &\sup_{f: \calS \times \calA \rightarrow \R}  \Big[ \E_{(s, a) \sim \mu^{\pi'}_\rho}[ f(s, a)]-  \\
    & \quad \log \left( \E_{(s, a) \sim \mu^{\pi}_\rho}[ \exp( f(s, a))] \right) \Big]. \nonumber
\end{align}
The \textit{log-expected-exp} in this equation makes the Donkser-Varadhan representation~\eqref{eq:donkser} more numerically stable than the variational one~\eqref{eq:dice} when working with KL divergences. Because of its genericity for $\phi$-divergences, we base the remainder of our exposition on~\eqref{eq:dice}. But it is straightforward to adapt the approach and algorithm to using~\eqref{eq:donkser} for better numerical stability when working with KL divergences specifically. Thus, in practice we will use the latter in our experiments with KL-based regularization, but not in the ones with $\chi^2$-based regularization.

\section{A PRACTICAL ALGORITHM USING ADVERSARIAL DIVERGENCE}

\begin{algorithm*}[h!]
\caption{\textsc{PPO-DICE}}\label{alg:main}

\begin{algorithmic}[1]
\State \textbf{Initialisation}: random initialize parameters  $\theta_1$ (policy), $\psi_1$ (discriminator) and $\omega_1$ (value function).
\For{i=1, \ldots}
\State Generate a batch of $M$ rollouts $\{s^{(j)}_1, a^{(j)}_1, r^{(j)}_1, s^{(j)}_{1}, \ldots, s^{(j)}_{T}, a^{(j)}_{T}, r^{(j)}_T, s^{(j)}_{T+1}\}_{j=1}^M$ by executing policy $\pi_{\theta_i}$ in the environment for $T$ steps.
\State Estimate Advantage function: $\hat{A}(s^{(j)}_t, a^{(j)}_t) = \sum_{t=1}^T (\gamma \lambda)^{t-1} (r^{(j)}_t + \gamma V_{\omega_i}(s^{(j)}_{t+1}) - V_{\omega_i}(s^{(j)}_t))$
 \State Compute target value $y^{(j)}_t = r^{(j)}_t + \gamma r^{(j)}_{t+1} + \ldots + \gamma^{T+1-t} V_{\omega_i}(s_{T+1})$
\State $\omega = \omega_i; \theta = \theta_i; \psi = \psi_i$
\For{epoch n=1, \ldots N}
   \For{iteration k=1, \ldots K}
       \State {\bf  \textcolor{gray!70!blue}{// Compute discriminator loss:}} 
    \State $
        \hat{L}_D(\psi, \theta) = \frac{1}{MT}\sum_{j=1}^M\sum_{t=1}^{T}  \phi^\star\left(g_\psi(s^{(j)}_t, a^{(j)}_t) - \gamma g_\psi(s^{(j)}_{t+1}, a'^{(j)}_{t+1})\right)- (1 - \gamma) g_\psi(s^{(j)}_1, a'^{(j)}_t)
    $
    where $a'^{(j)}_t \sim \pi_\theta(\cdot \mid s^{(j)}_1),  a'^{(j)}_{t+1} \sim \pi_\theta(\cdot \mid s^{(j)}_{t+1})$.
    \State {\bf  \textcolor{gray!70!blue}{// Update discriminator parameters:}} (using learning rate $c_\psi \eta$)
    \State $
         \psi  \leftarrow \psi - c_\psi\eta \nabla_{\psi} \hat{L}_D(\psi, \theta);$
    \EndFor
    \State {\bf  \textcolor{gray!70!blue}{// Compute value loss:}}
    \State $
        \hat{L}_V(\omega) =  \frac{1}{MT}\sum_{j=1}^M\sum_{t=1}^{T}\left(V_\omega(s_t^{(j)}) - y^{(j)}_t\right)^2 $
    
    \State {\bf  \textcolor{gray!70!blue}{// Compute PPO clipped loss:}}
    \State $
        \hat{L}^{\mathrm{clip}}(\theta)  = 
        \frac{1}{MT}\sum_{j=1}^M\sum_{t=1}^{T} \min\big\{ \hat{A}(s^{(j)}_t, a^{(j)}_t) \kappa_{\pi_\theta /\pi_{\theta_i}}(s^{(j)}_t, a^{(j)}_t), \hat{A}(s^{(j)}_t, a^{(j)}_t) \mathrm{clip}(\kappa_{\pi_\theta /\pi_{\theta_i}}(s^{(j)}_t, a^{(j)}_t), 1-\epsilon, 1+\epsilon)\big\}
    $
    \State {\bf  \textcolor{gray!70!blue}{// Update parameters:}} (using learning rate $\eta$)
    
    \State{$
        \omega \leftarrow \omega - \eta \nabla_{\omega} \hat{L}_V(\omega); 
        $}
    \State{$          \theta \leftarrow \theta + \eta \nabla_{\theta} (\hat{L}^{\mathrm{clip}}(\theta) + \lambda \cdot \hat{L}_D(\psi,\theta))
    $ (if reparametrization trick applicable, else gradient step on Eq.~\ref{eq:empirical-score-function-objective})}
\EndFor
\State $\omega_{i+1} = \omega; \theta_{i+1} = \theta; \psi_{i+1} = \psi$
\EndFor

\end{algorithmic}
\end{algorithm*}

We now turn these insights into a practical algorithm.
The lower bounds in lemma ~\ref{lemma:lower_bound_perf}, suggest using a regularized PPO objective\footnote{\label{footnote:clip} Both regularized  $L^\mathrm{clip}_{\pi_i}$ and $L_{\pi_i}$ are lower bounds on policy performance in Lemma \ref{lemma:lower_bound_perf}. We use $L^\mathrm{clip}_{\pi_i}$ rather than $L_{\pi_i}$ because we expect it to work better as the clipping already provides some constraint on action probabilities. Also this will allow a more direct empirical assessment of what the regularization brings compared to vanilla PPO.} : 
$L^{\mathrm{clip}}_{\pi}(\pi') - \lambda D_{\TV}(d^{\pi'}_\rho \| d^{\pi}_\rho)$, where $\lambda$ is a regularization coefficient. If in place of the total variation we use the off-policy formulation of $\phi$-divergence $D_\phi(\mu^{\pi'}_\rho \|  \mu^{\pi}_\rho)$ as detailed in Equation \eqref{eq:dice}, our main optimization objective can be expressed as the following min-max problem:
\begin{align}
     \max_{\pi'} & \min_{g: \calS \times \calA \rightarrow \R} L^{\mathrm{clip}}_{\pi_i}(\pi') - \lambda \Big( (1-\gamma) \E_{s \sim \rho, a \sim \pi'}[ g(s, a)]- \nonumber \\
    & \E_{(s, a) \sim \mu^{\pi_i}_\rho}\left[ \phi^\star  \left((g - \gamma \P^{\pi'}g)(s,a)\right)\right] \Big),
    \label{eq:main_obj}
\end{align}
When the inner minimization over $g$ is fully optimized, it is straightforward to show -- using the score function estimator -- that the gradient of this objective with respect to $\pi$ is (proof is provided in appendix):
\begin{align}
\label{eq:score function estimator}
& \nabla_{\pi'} L^{\mathrm{clip}}_{\pi_i}({\pi'}) - \lambda \Big( (1-\gamma) \E_{\substack {s \sim \rho \\ a \sim \pi'}}[ g(s, a) \nabla_{\pi'} \log\pi'(a \mid s)] \nonumber \\
& + \gamma \E_{(s, a) \sim \mu^{\pi_i}_\rho}\big[ \frac{\partial \phi^{\star}}{\partial t} \left((g - \gamma \P^{\pi'}g)(s,a)\right) \\
& \E_{s' \sim \P(\cdot \mid s, a), a' \sim \pi'(\cdot \mid s')} \left[ g(s', a') \nabla_{\pi'} \log\pi'(a' \mid s')\right]\big] \Big). \nonumber
\end{align}
Furthermore, we can use the reparametrization trick if the policy $\pi$ is parametrized by a Gaussian, which is usually the case in continuous control tasks. We call the resulting new algorithm PPO-DICE, (detailed in Algorithm~\ref{alg:main}), as it uses the clipped loss of PPO and leverages the DIstribution Correction Estimation idea from \citet{nachum2019dualdice}.

In the min-max objective~\eqref{eq:main_obj},
$g$ plays the role of a discriminator, as in Generative Adversarial Networks (GAN)~\citep{goodfellow2014generative}. The policy $\pi'$ plays the role of a generator, and it should balance between increasing the likelihood of actions with large advantage versus inducing a state-action distribution that is close to the one of $\pi_i$. 

As shown in Algorithm ~\ref{alg:main}, both policy and discriminator are parametrized by neural networks $\pi_\theta$ and $g_\psi$ respectively. We estimate the objective~\eqref{eq:main_obj} with samples from $\pi_i = \pi_{\theta_i}$ as follows. At a given iteration $i$, we generate a batch of $M$ rollouts $\{s^{(j)}_1, a^{(j)}_1, r^{(j)}_1, s^{(j)}_{1}, \ldots, s^{(j)}_{T}, a^{(j)}_{T}, r^{(j)}_T, s^{(j)}_{T+1}\}_{j=1}^M$ by executing the policy $\pi_i$ in the environment for $T$ steps. Similarly to the PPO procedure, we learn a value function $V_\omega$ by updating its parameters $\omega$ with gradient descent steps, optimizing the following squared error loss:
\begin{equation}
    \hat{L}_V(\omega) = \frac{1}{MT}\sum_{j=1}^M\sum_{t=1}^{T}\left(V_\omega(s_t^{(j)}) - y^{(j)}_t\right)^2,
\end{equation}
where $y^{(j)}_t = r^{(j)}_t + \gamma r^{(j)}_{t+1} + \ldots + \gamma^{T+1-t}V_\omega(s_{T+1})$. Then, to estimate the advantage, we use the truncated generalized advantage estimate
\begin{equation}
    \hat{A}(s^{(j)}_t, a^{(j)}_t) = \sum_{t=1}^T (\gamma \lambda)^{t-1} (r^{(j)}_t + \gamma V_\omega(s^{(j)}_{t+1}) - V_\omega(s^{(j)}_t)). 
\end{equation}
This advantage estimate is used to compute an estimate of $L^{\mathrm{clip}}_{\pi_i}$ given by:
\begin{align}
    & \hat{L}^{\mathrm{clip}}(\theta) =  \\
    &\quad  
    \frac{1}{MT}\sum_{j=1}^M\sum_{t=1}^{T} \min\Big\{ \hat{A}(s^{(j)}_t, a^{(j)}_t) \kappa_{\pi_\theta /\pi_{i}}(s^{(j)}_t, a^{(j)}_t), \nonumber \\
    & \quad \hat{A}(s^{(j)}_t, a^{(j)}_t) \cdot \mathrm{clip}(\kappa_{\pi_\theta /\pi_{i}}(s^{(j)}_t, a^{(j)}_t), 1-\epsilon, 1+\epsilon)\Big\} \nonumber
\end{align}
The parameters $\psi$ of the discriminator are learned by gradient descent on the following empirical version of the regularization term in the min-max objective \eqref{eq:main_obj}
\begin{align}
    \label{eq:empirical_reg}
     \hat{L}_D(\psi, \theta) & = \frac{-1}{MT}\sum_{j=1}^M\sum_{t=1}^{T} (1 - \gamma) g_\psi(s^{(j)}_1, a'^{(j)}_t) \\ 
     & \quad - \phi^\star\left(g_\psi(s^{(j)}_t, a^{(j)}_t) - \gamma g_\psi(s^{(j)}_{t+1}, a'^{(j)}_{t+1})\right), \nonumber
\end{align}
where $a'^{(j)}_t \sim \pi_\theta(\cdot \mid s^{(j)}_1)$ and $ a'^{(j)}_{t+1} \sim \pi_\theta(\cdot \mid s^{(j)}_{t+1})$.

If the reparametrization trick is applicable (which is almost always the case for continuous control tasks), the parameters $\theta$ of the policy are updated via gradient ascent on the objective $\hat{L}^{\mathrm{clip}}(\theta) + \lambda \hat{L}_D(\psi, \theta)$ as we can backpropagate gradient though the action sampling while computing $\hat{L}_D(\psi, \theta)$ in Equation~\eqref{eq:empirical_reg}. Otherwise, $\theta$ are updated via gradient ascent on the following objective:
\begin{align}
     & \quad \hat{L}^{\mathrm{clip}}(\theta) - \nonumber \\ 
     & \quad \frac{\lambda}{MT}\sum_{j=1}^M\sum_{t=1}^{T} (1 - \gamma) g_\psi(s^{(j)}_1, a'^{(j)}_t) \log \pi_\theta(a'^{(j)}_t \mid s^{(j)}_1) \nonumber \\ 
     & \quad + \gamma \frac{\partial \phi^{\star}}{\partial t}\left(g_\psi(s^{(j)}_t, a^{(j)}_t) - \gamma g_\psi(s^{(j)}_{t+1}, a'^{(j)}_{t+1})\right) \nonumber \\
     & \quad \cdot  g_\psi(s^{(j)}_{t+1}, a'^{(j)}_{t+1}) \log \pi_\theta(a'^{(j)}_{t+1}) \mid s^{(j)}_{t+1})  \label{eq:empirical-score-function-objective}
\end{align}
Note that the gradient of this equation with respect to $\theta$ corresponds to an empirical estimate of the score function estimator we provided in Equation~\ref{eq:score function estimator}.

We train the value function, policy, and discriminator for $N$ epochs using $M$ rollouts of the policy $\pi_i$.
We can either alternate between updating the policy and the discriminator, or update $g_\psi$ for a few steps $M$ before updating the policy. We found that the latter worked better in practice, likely due to the fact that the target distribution $\mu^{\pi_i}_\rho$ changes with every iteration $i$. We also found that increasing the learning rate of the discriminator by a multiplicative factor $c_\psi$ of the learning rate for the policy and value function $\eta$ improved performance.
\paragraph{Choice of divergence:} 
The algorithmic approach we just described is valid with any choice of $\phi$-divergence for measuring the discrepancy between state-visitation distributions. It remains to choose an appropriate one.
While Lemma~\ref{lemma:lower_bound_perf} advocates the use of total variation distance ($\phi(t) = |t-1|$), it is notoriously hard to train high dimensional distributions using this divergence  (see~\citet{kodali2017convergence} for example). Moreover, the convex conjugate of $\phi(t) = |t-1|$ is $\phi^\star(t) = t$ if $ |t| \leq \frac{1}{2}$ and $\phi^\star(t) = \infty$ otherwise. This would imply the need to introduce an extra constraint $\|g - \P^\pi g\|_\infty \leq \frac{1}{2}$ in the formulation~\eqref{eq:dice}, which may be hard to optimize.  

Therefore, we will instead use  the KL divergence ($\phi(t) = t \log(t), \phi^\star (t) = \exp(t-1)$). This is still a well justified choice as we know that $D_{\TV}(\mu^{\pi'}_\rho \| \mu^\pi_\rho) \leq \sqrt{\frac{1}{2} D_{\KL}(\mu^{\pi'}_\rho \| \mu^\pi_\rho)}$ thanks to Pinsker's inequality. We will also try $\chi^2$-divergence ($\phi(t)=(t-1)^2$) that yields a squared regularization term.
\section{RELATED WORK}
Constraining policy updates, in order to minimize the information loss due to policy improvement, has been an active area of investigation.~\citet{kakade2002approximately} originally introduce CPI by maximizing a lower bound on the policy improvement and relaxing the greedification step through a mixture of successive policies.~\citet{pirotta2013safe} build on~\citet{kakade2002approximately} refine the lower bounds and introduce a new mixture scheme. Moreover, CPI inspired some popular Deep RL algorithms such as  TRPO~\citep{schulman2015trust} and PPO~\citep{schulman2015trust}, Deep CPI~\citep{vieillard2019deep} and MPO~\citep{AbdolmalekiSTMH18}. The latter uses similar updates to TRPO/PPO in the parametric version of its E-step. So, our method can be incorporated to it. 

Our work is related to regularized MDP literature~\citep{neu2017unified,geist2019theory}. Shannon Entropic regularization is used in value iteration scheme~\citep{haarnoja2017reinforcement, dai2018sbeed} and in policy iteration schemes~\citep{haarnoja2018soft}. Note that all the mentioned works employ regularization on the action probabilities.
Recently,~\citet{wang2019divergence}
introduce divergence-augmented policy optimization where they penalize the policy update by a Bregman divergence on the state visitation distributions, motivated the mirror descent method. While their framework seems general, it doesn't include the divergences we employ in our algorithm. In fact, their method enables the use of the \emph{conditional} KL divergence between state-action visitations distribution defined by $\int \mu_\rho^\pi(s, a) \log \frac{\pi(a \mid s)}{\pi'(a \mid a)}$ and not the KL divergence $\int \mu_\rho^\pi(s, a) \log \frac{\mu_\rho^\pi(s, a)}{\mu_\rho^{\pi'}(s, a)}$. Note the action probabilities ratio inside the $\log$ in the conditional KL divergence allows them to use the policy gradient theorem, a key ingredient in their framework, which cannot be done for the KL divergence.

 Our work builds on recent off-policy approaches: DualDICE~\citep{nachum2019dualdice} for policy evaluation and ValueDICE~\citep{kostrikov2019imitation} for imitation learning. Both use the off-policy formulation of KL divergence. The former uses the formulation to estimate the ratio of the state
visitation distributions under the target and behavior policies. Whereas, the latter learns a policy by minimizing the divergence. 

The closest related work is the recently proposed AlgaeDICE~\citep{nachum2019algaedice} for off-policy policy optimization. They use the divergence between state-action visitation distribution induced by $\pi$ and a behavior distribution, motivated by similar techniques in~\citet{nachum2019dualdice}. However, they incorporate the regularization to the dual form of policy performance $J(\pi) = \E_{(s, a) \sim \mu^\pi_\rho}[r(s, a)]$ whereas we consider a surrogate objective (lower bound on the policy performance). Moreover, our method is online off-policy in that we collect data with each policy found in
the optimization procedure, but also use previous data to improve stability. Whereas, their algorithm is designed to learn a policy from a fixed dataset collected by behaviour policies. Further comparison with AlgaeDICE is provided in appendix.

\section{EXPERIMENTS AND RESULTS}
We use the PPO implementation by \cite{pytorchrl} as a baseline and modify it to implement our proposed PPO-DICE algorithm. We run experiments on a randomly selected subset of environments in the Atari suite \citep{ale} for high-dimensional observations and discrete action spaces, as well as on the OpenAI Gym \citep{openaigym} MuJoCo environments, which have continuous state-action spaces. All shared hyperparameters are set at the same values for both methods, and we use  the hyperparameter values recommended by \cite{pytorchrl} for each set of environments, Atari and MuJoCo~\footnote{Code: https://github.com/facebookresearch/ppo-dice}.

\subsection{IMPORTANT ASPECTS OF PPO-DICE}

\subsubsection{Choice of Divergence}
\begin{figure}[t]
    \centering
    \includegraphics[width=0.2\textwidth]{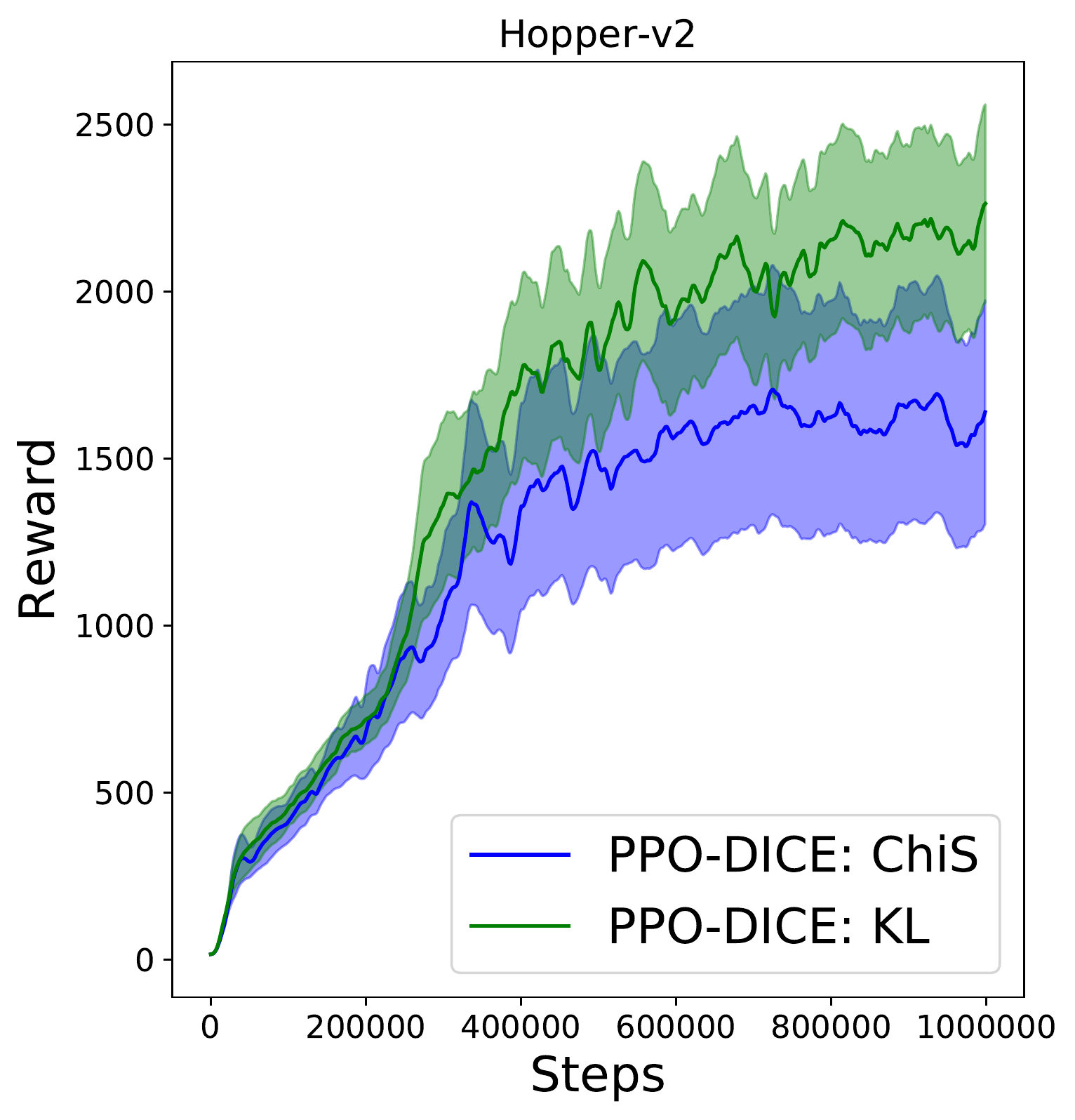} \hspace{10pt}
    \includegraphics[width=0.2\textwidth]{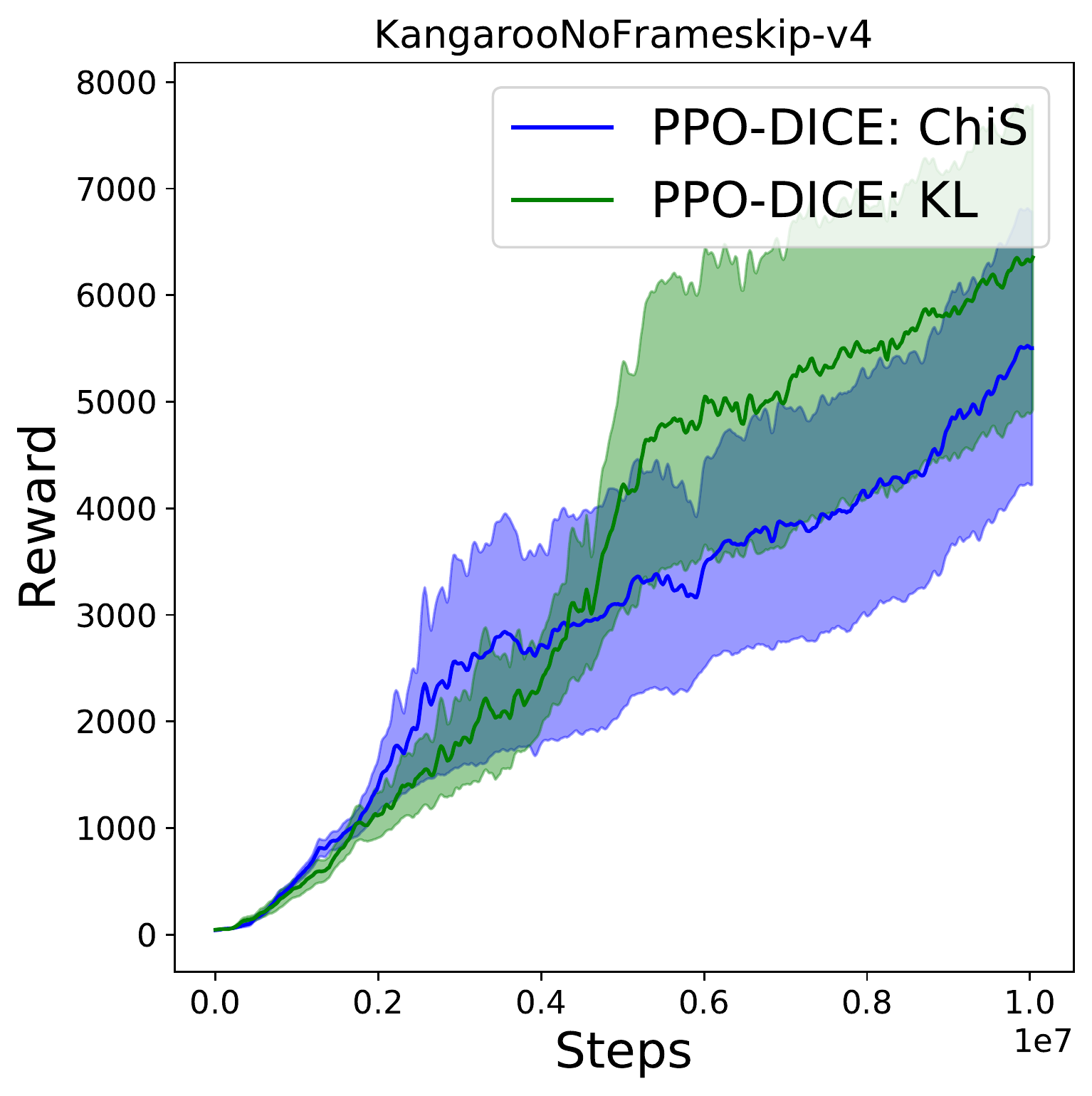}
    \caption{Comparison of $\chi^2$ and KL divergences for PPO-DICE for two randomly selected environments in OpenAI Gym MuJoCo and Atari, respectively. We see that KL performs better than $\chi^2$ in both settings. Performance plotted across 10 seeds with 1 standard error shaded.}
    \label{fig:hopper_chis}
\end{figure}
We conducted an initial set of experiments to compare two different choices of divergences, KL and $\chi^2$, for the regularization term of PPO-DICE. \cref{fig:hopper_chis} shows training curves for  one continuous action and one discrete action environment. There, as in the other environments in which we run this comparison,  KL consistently performed better than $\chi^2$. 
We thus opted to use KL divergence in all subsequent experiments.

\subsubsection{Effect of Varying $\lambda$}
\label{sec:lambda}
\begin{figure}[h]
    \centering
    \includegraphics[width=0.27\textwidth]{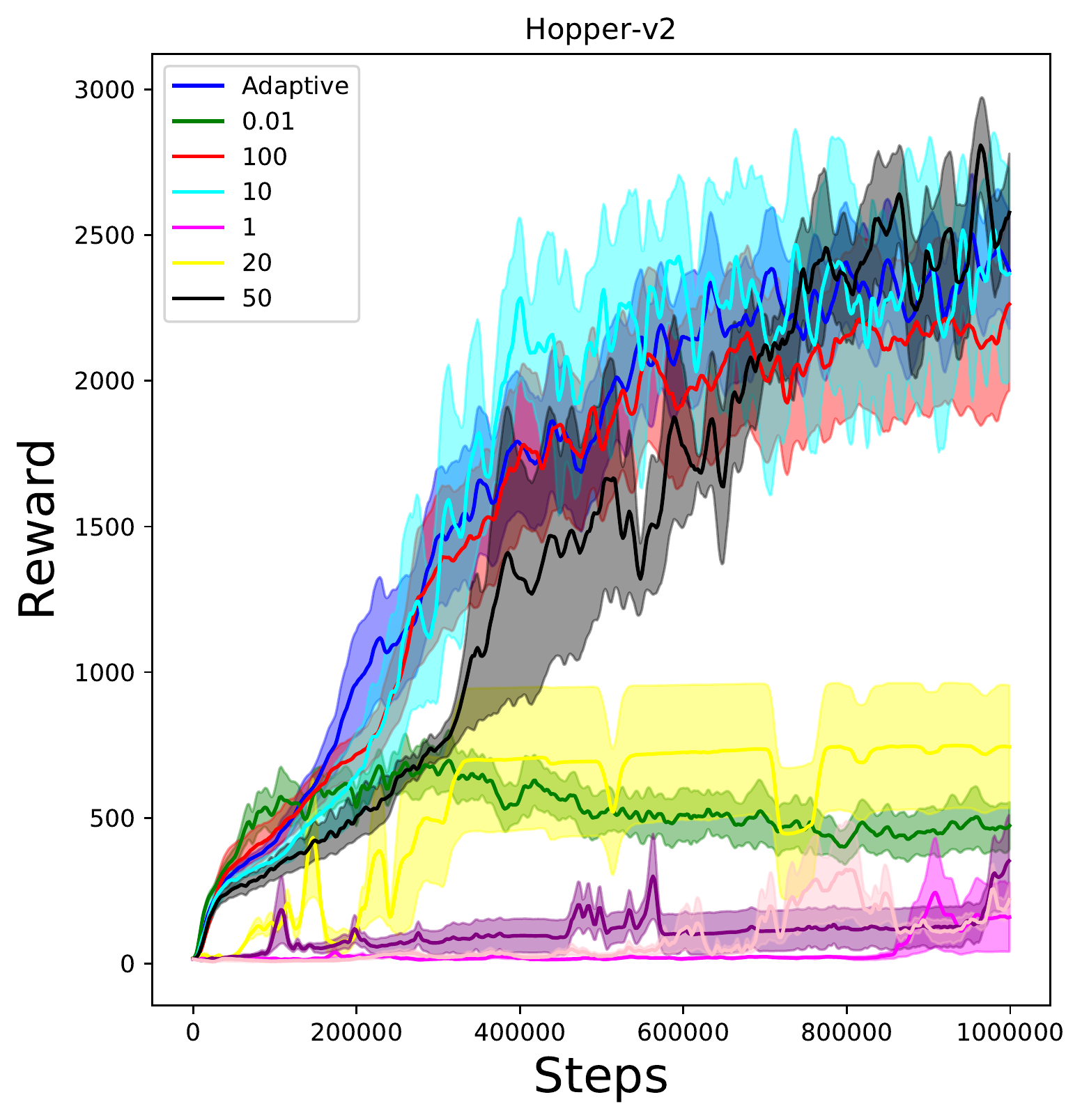}
    \caption{Varying $\lambda$ in \texttt{Hopper\_v2}, 10 seeds, 1 standard error shaded. PPO-DICE is somewhat sensitive to $\lambda$ value, but the theoretically-motivated adaptive version works well.}
    \label{fig:hopper_lambda}
\end{figure}
Next we wanted to evaluate the sensitivity of our method to the $\lambda$ parameter that controls the strength of the regularization.  We examine in \cref{fig:hopper_lambda} the performance of PPO-DICE when varying $\lambda$. There is a fairly narrow band for \texttt{Hopper-v2} that performs well, between $0.01$ and $1$. Theory indicates that the proper value for $\lambda$ is the maximum of the absolute value of the advantages (see Lemma \ref{lemma:lower_bound_perf}). This prompted us to implement an adaptive approach, where we compute the 90th percentile of advantages within the batch (for stability), which % we show 
we found performed well across environments. To avoid introducing an additional hyperparameter by tuning $\lambda$, we use the adaptive method for subsequent experiments.
\begin{figure}[h!]
    \centering
    \includegraphics[width=0.27\textwidth]{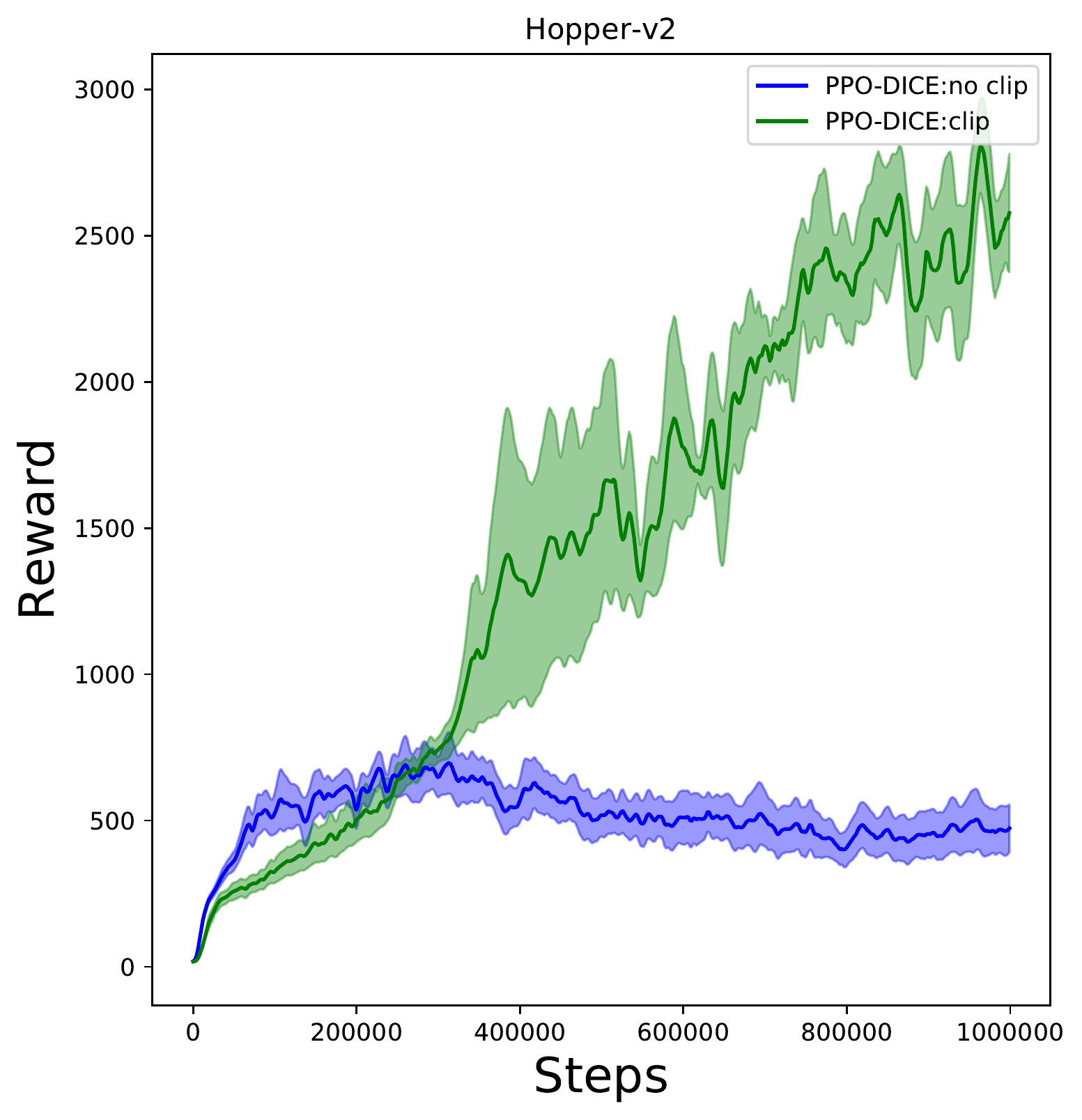}
    \caption{Comparison of PPO-DICE with clipped loss $L^\text{clip}$ and without $L$. We see that clipping the action loss is crucial for good performance.}
    \label{fig:hopper_noclip}
\end{figure}

\begin{table*}[h!]
% \small
\vspace{5px} 
    \centering
    \vspace{-10pt}
    \begin{tabular}{l|ll}
    \toprule
    Game & PPO & PPO-DICE \\
    \midrule
AirRaid & $ 4305.0 \pm 638.15 $ & $ \textcolor{blue}{\mathbf{5217.5 \pm 769.19 }}$ \\
Asterix & $ 4300.0 \pm 169.31 $ & $ \textcolor{blue}{\mathbf{6200.0 \pm 754.10 }}$ \\
Asteroids & $ 1511.0 \pm 125.03 $ & $\textcolor{blue}{\mathbf{ 1653.0 \pm 112.20 }}$ \\
Atlantis & $ 2120400.0 \pm 471609.93 $ & $\textcolor{blue}{\mathbf{ 3447433.33 \pm 100105.82}} $ \\
BankHeist & $ 1247.0 \pm 21.36 $ & $ \textcolor{blue}{\mathbf{1273.33 \pm 7.89 }}$ \\
BattleZone & $\textcolor{blue}{\mathbf{ 29000.0 \pm 2620.43}} $ & $ 19000.0 \pm 2463.06 $ \\
Carnival & $ 3243.33 \pm 369.51 $ & $ 3080.0 \pm 189.81 $ \\
ChopperCommand & $ 566.67 \pm 14.91 $ & $ \textcolor{blue}{\mathbf{900.0 \pm 77.46}} $ \\
DoubleDunk & $ -6.0 \pm 1.62 $ & $ \textcolor{blue}{\mathbf{-4.0 \pm 1.26}} $ \\
Enduro & $ 1129.9 \pm 73.18 $ & $ \textcolor{blue}{\mathbf{1308.33 \pm 120.09}} $ \\
Freeway & $ 32.33 \pm 0.15 $ & $ 32.0 \pm 0.00 $ \\
Frostbite & $ \textcolor{blue}{\mathbf{639.0 \pm 334.28}} $ & $ 296.67 \pm 5.96 $ \\
Gopher & $ 1388.0 \pm 387.65 $ & $ 1414.0 \pm 417.84 $ \\
Kangaroo & $ 4060.0 \pm 539.30 $ & $\textcolor{blue}{\mathbf{ 6650.0 \pm 1558.16 }}$ \\
Phoenix & $ \textcolor{blue}{\mathbf{12614.0 \pm 621.71}} $ & $ 11676.67 \pm 588.24 $ \\
Robotank & $ 7.8 \pm 1.33 $ & $\textcolor{blue}{\mathbf{ 12.1 \pm 2.91}} $ \\
Seaquest & $ 1198.0 \pm 128.82 $ & $ 1300.0 \pm 123.97 $ \\
TimePilot & $ 5070.0 \pm 580.53 $ & $ \textcolor{blue}{\mathbf{7000.0 \pm 562.32}} $ \\
Zaxxon & $ \textcolor{blue}{\mathbf{7110.0 \pm 841.60 }}$ & $ 6130.0 \pm 1112.48 $ \\
     \bottomrule
     \end{tabular}
     \caption{Mean final reward and 1 standard error intervals across 10 seeds for Atari games evaluated at 10M steps.}
     \label{tab:atari}
\end{table*}

\begin{figure*}[h!]
    \centering
    \includegraphics[width=0.195\textwidth]{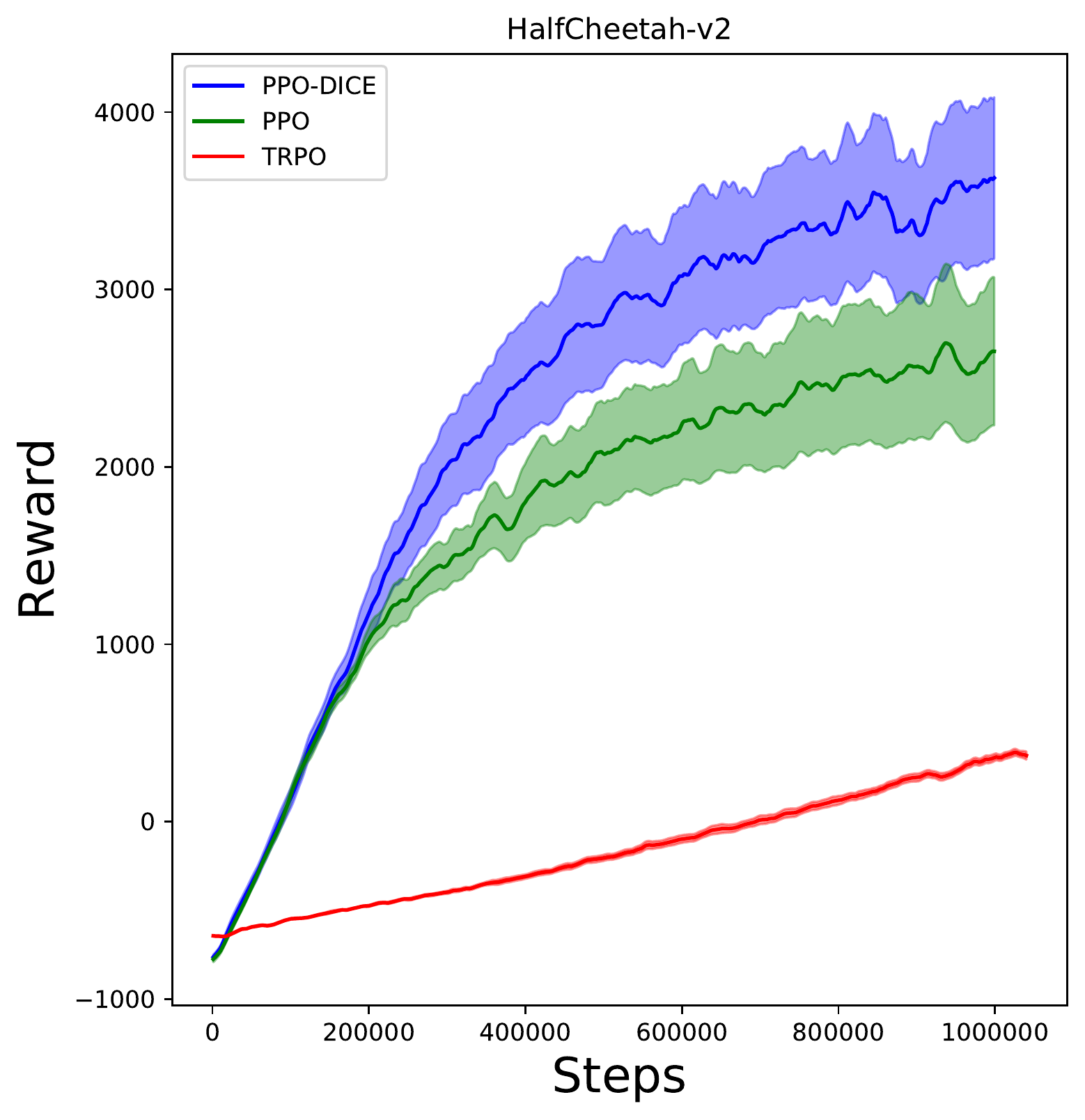}
    \includegraphics[width=0.195\textwidth]{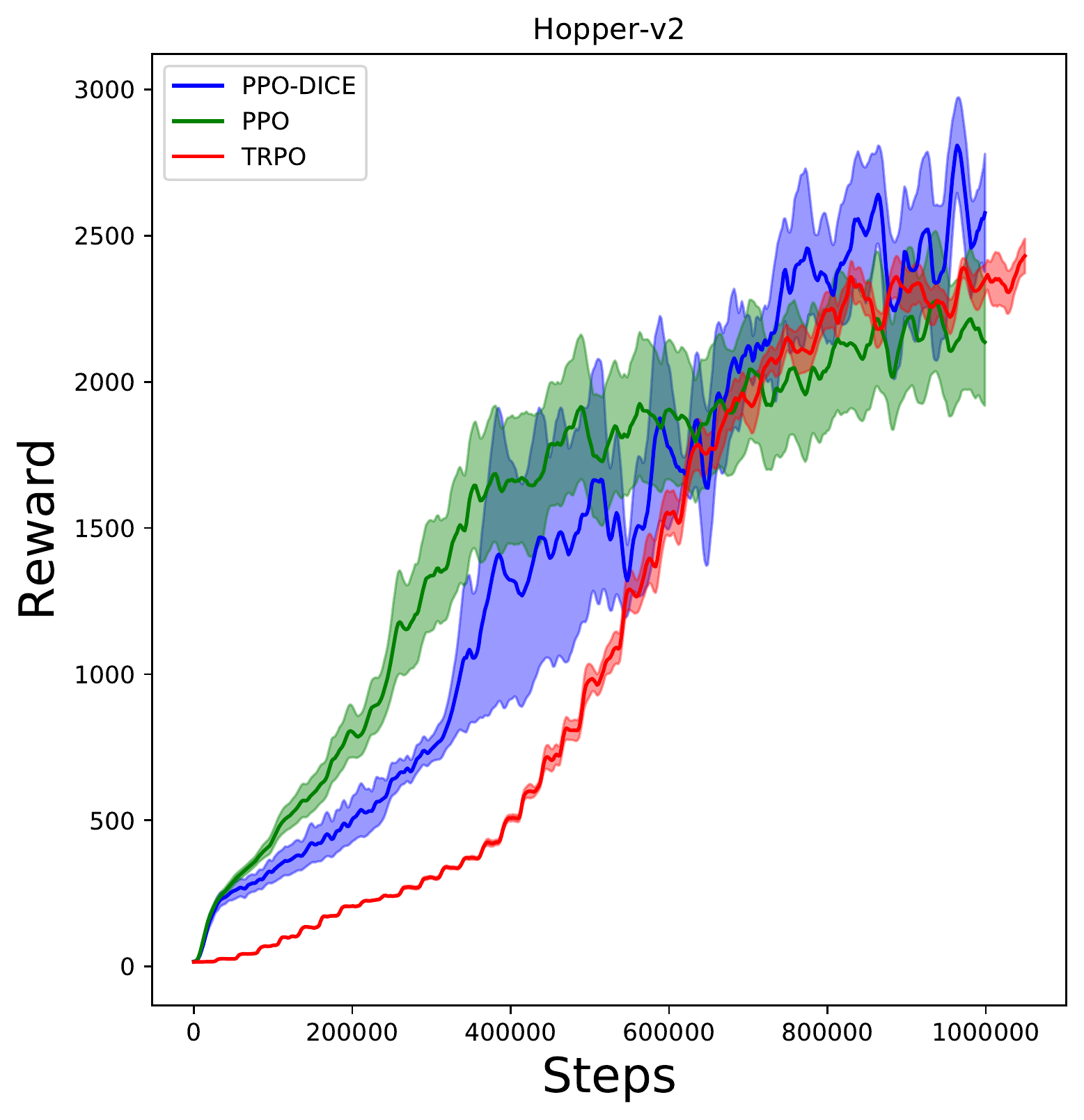}
    \includegraphics[width=0.195\textwidth]{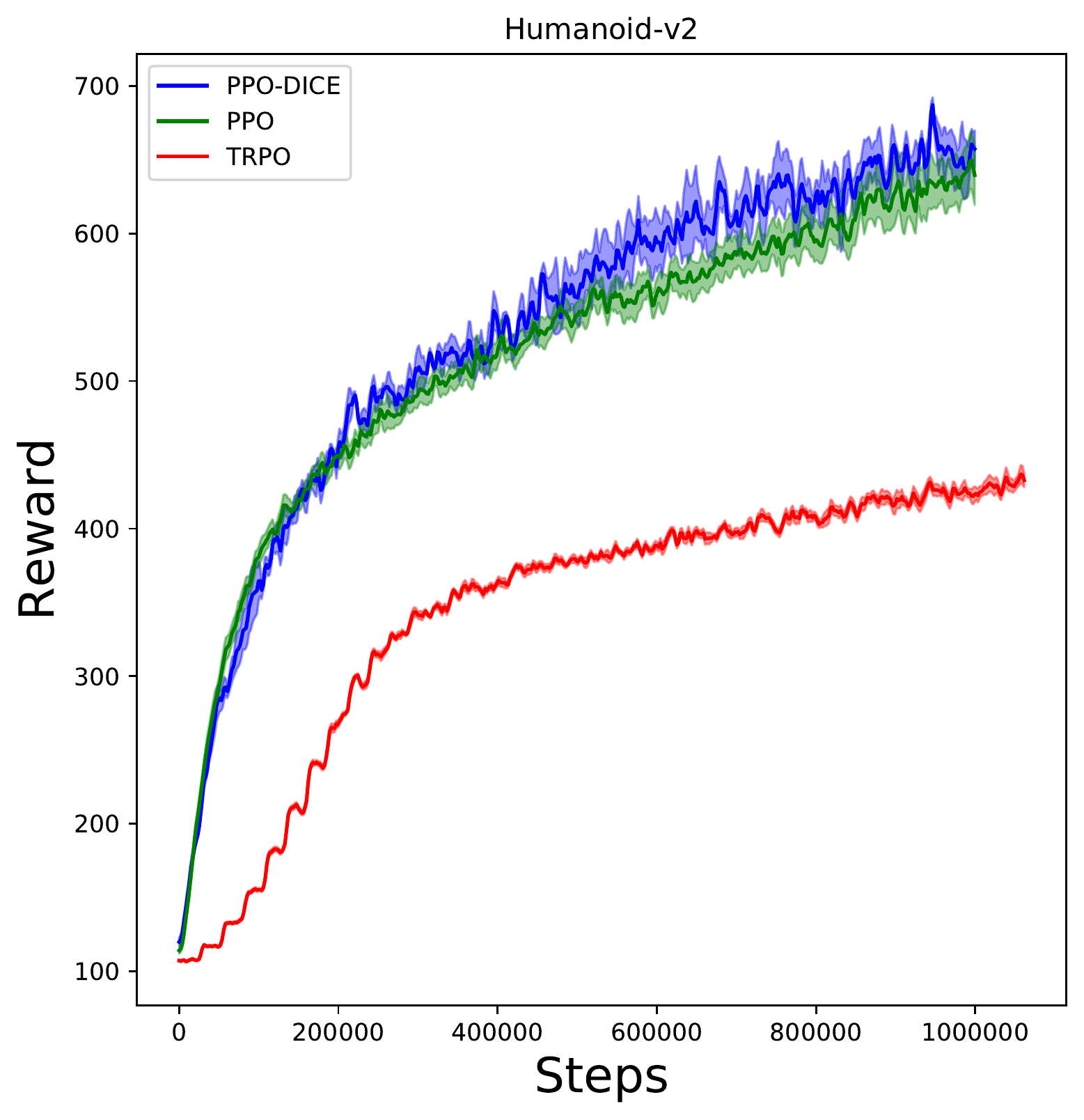}
    \includegraphics[width=0.195\textwidth]{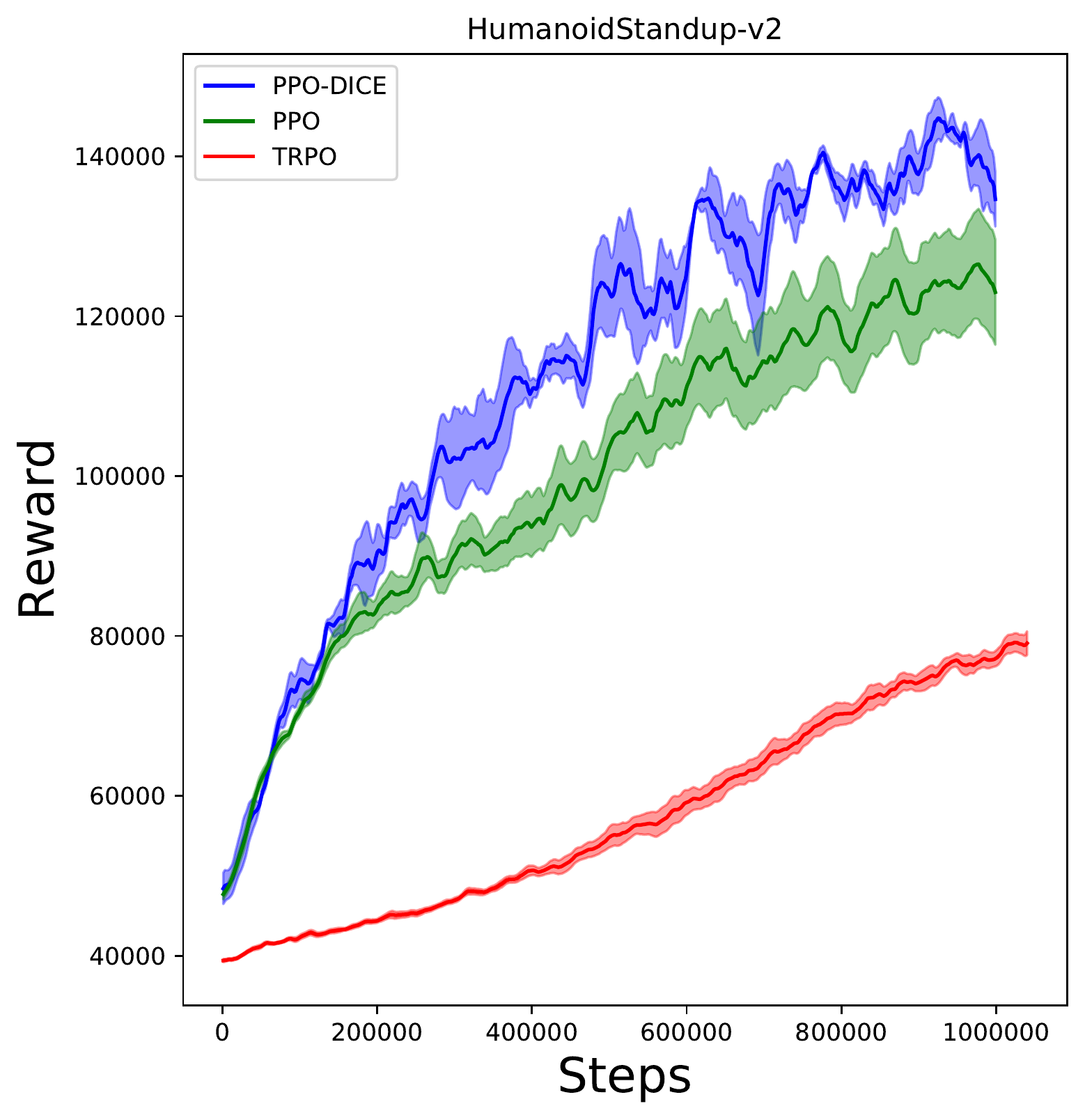}
    \includegraphics[width=0.195\textwidth]{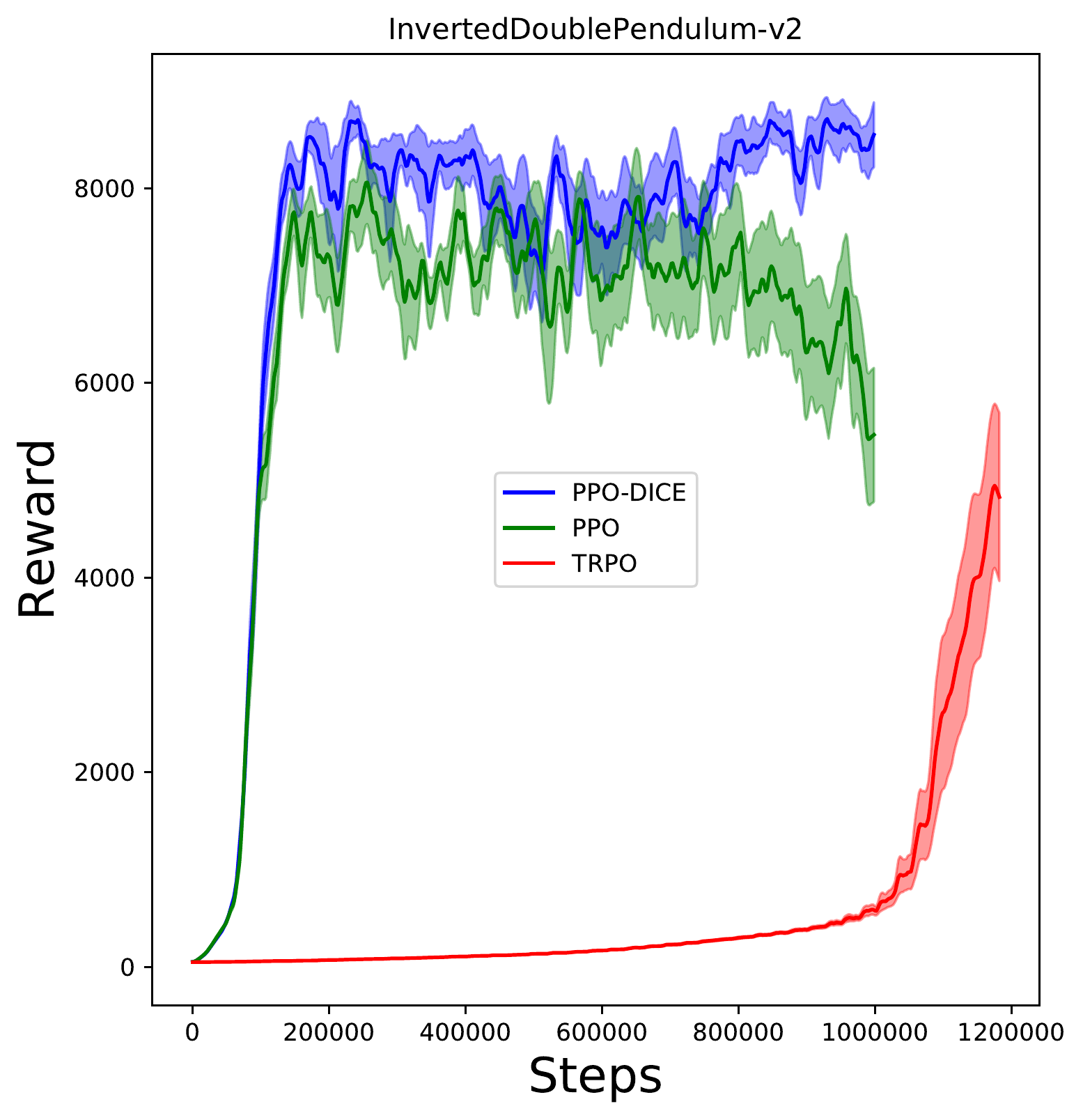}
    \caption{Results from OpenAI Gym MuJoCo suite in more complex domains, with 10 seeds and 1 standard error shaded.  Results on the full suite of environments can be found in \cref{app:mujoco_results}.
    \label{fig:mujoco_results}}
\end{figure*}

\subsubsection{Importance of Clipping the Action Loss}

We earlier mentioned (see \cref{footnote:clip}) two possible forms of our regularized objective: one with clipped action loss $L^{\text{clip}}$ and one without $L$. Clipping the action loss was an extra regularizing measure proposed in PPO~\citep{schulman2017proximal}. For our algorithm also, we hypothesized that it would be important for providing additional constraints on the policy update to stay within the trust region. \cref{fig:hopper_noclip} confirms this empirically: we see the effect on our method of clipping the action loss versus keeping it unclipped. Initially, not having the additional regularization allows it to learn faster, but it soon crashes, showing the need for clipping to reduce variance in the policy update.

\subsection{RESULTS ON ATARI}

Given our above observations we settled on using a KL-regularized $L^\mathrm{clip}$, with the adaptive method for $\lambda$ that we explained Section \ref{sec:lambda}. We run PPO-DICE on randomly selected environments from Atari. We tuned two additional hyperparameters, the learning rate for the discriminator and the number of discriminator optimization steps per policy optimization step. We found that $K=5$ discriminator optimization steps per policy optimization step performed well. Fewer steps showed worse performance because the discriminator was not updating quickly enough, while more optimization steps introduced instability from the discriminator overfitting to the current batch. We also found that increasing the discriminator learning rate to be $c_\psi=10\times$ the policy learning rate helped most environments. We used the same hyperparameters across all environments. 
Results are shown in \cref{tab:atari}.  
We see that PPO-DICE significantly outperforms PPO on a majority of Atari environments. See \cref{app:atari_results}  for training curves and hyperparameters. 

\subsection{RESULTS ON OpenAI Gym MuJoCo}

For the OpenAI Gym MuJoCo suite, we also used $K=5$ discriminator optimization steps per policy optimization step, and $c_\psi=10\times$ learning rate for the discriminator in all environments.
We selected 5 of the more difficult environments to showcase in the main paper (\cref{fig:mujoco_results}), but additional results on the full suite and all hyperparameters used can be found in \cref{app:mujoco_results}. We again see improvement in performance in the majority of environments with PPO-DICE compared to PPO and TRPO. 

\section{CONCLUSION}
In this work, we have argued that using the action probabilities to constrain the policy update is a suboptimal approximation to controlling the state visitation distribution shift.
We then demonstrate that using the recently proposed DIstribution Correction Estimation idea~\citep{nachum2019dualdice}, we can directly compute the divergence between the state-action visitation distributions of successive policies and use that to regularize the policy optimization objective instead. Through carefully designed experiments, we have shown that our method beats PPO in most environments in Atari~\citep{ale} and OpenAI Gym MuJoCo~\citep{openaigym} benchmarks. 
\section{Acknowledgements}
We would like to thank Ofir Nachum and Ilya Kostrikov for their helpful feedback and advice during discussions at the early stage of the project.
% \newpage
\bibliographystyle{apalike}
\bibliography{lib}

\appendix
\onecolumn
\section{Omitted Proofs}
\subsection{Proof of Lemma~\ref{lemma:lower_bound_perf}}
According to performance difference lemma~\ref{lemma:performance diff}, we have
\begin{align*}
    J(\pi') & = J(\pi) + \E_{s \sim d_\rho^{\pi'}} \E_{a \sim \pi'(\cdot \mid s)} \left[ A^{\pi}(s, a) \right] \\
    & = J(\pi) + \left ( \E_{s \sim d_\rho^{\pi}} \E_{a \sim \pi'(\cdot \mid s)} \left[ A^{\pi}(s, a) \right] + \int_{s \in \calS}  \E_{a \sim \pi'(\cdot \mid s)} \left[ A^{\pi}(s, a) \right] (d^{\pi'}_\rho(s) - d^{\pi}_\rho(s)) ds \right) \\
    & \geq J(\pi) + \left ( \E_{s \sim d_\rho^{\pi}} \E_{a \sim \pi'(\cdot \mid s)} \left[ A^{\pi}(s, a) \right] - \int_{s \in \calS} | \E_{a \sim \pi(\cdot \mid s)} \left[ A^{\pi}(s, a) \right]| \cdot | d^{\pi'}_\rho(s) - d^{\pi}_\rho(s)|  ds \right) \\
    & \geq J(\pi) +  \left ( \E_{s \sim d_\rho^{\pi}} \E_{a \sim \pi'(\cdot \mid s)} \left[ A^{\pi}(s, a) \right] - \epsilon^{\pi} \int_{s \in \calS} | d^{\pi'}_\rho(s) - d^{\pi}_\rho(s)|  ds \right) \\
    & \geq J(\pi) +  \left ( \E_{s \sim d_\rho^{\pi}} \E_{a \sim \pi'(\cdot \mid s)} \left[ A^{\pi}(s, a) \right] - \epsilon^{\pi} D_{\TV}( d^{\pi'}_\rho\| d^{\pi}_\rho ) \right) \\
    & = L_{\pi}(\pi') - \epsilon^{\pi} D_{\TV}( d^{\pi'}_\rho\| d^{\pi}_\rho )
\end{align*}

where $\epsilon^{\pi} = \max_s | \E_{a \sim \pi'(\cdot \mid s)} \left[ A^{\pi}(s, a) \right]|$ and $D_{\TV}$ is total variation distance. The first inequality follows from Cauchy-Schwartz inequality.

\subsection{Score Function Estimator of the gradient with respect to the policy}

\begin{align*}
    \nabla_{\pi'} \E_{\substack {s \sim \rho \\ a \sim \pi'}}[ g(s, a) ] = \nabla_{\pi'} \int g(s, a) \rho(s) \pi'(a \mid s) = \int g(s, a) \rho(s)  \nabla_{\pi'} \pi'(a \mid s) = \E_{\substack {s \sim \rho \\ a \sim \pi'}}[ g(s, a) \nabla_{\pi'} \log\pi'(a \mid s)]
\end{align*}

\begin{align*}
    \nabla_{\pi'} \E_{(s, a) \sim \mu^{\pi_i}_\rho}\big[  \phi^{\star} \left((g - \gamma \P^{\pi'}g)(s,a)\right) \big] &=  \E_{(s, a) \sim \mu^{\pi_i}_\rho}\big[ \nabla_{\pi'} \phi^{\star} \left((g - \gamma \P^{\pi'}g)(s,a)\right) \big] \\
    & = \E_{(s, a) \sim \mu^{\pi_i}_\rho}\big[  \frac{\partial \phi^{\star}}{\partial t} \left((g - \gamma \P^{\pi'}g)(s,a)\right) \nabla_{\pi'} (g - \gamma \P^{\pi'}g) \big] \\
    & = -\gamma \E_{(s, a) \sim \mu^{\pi_i}_\rho}\big[ \frac{\partial \phi^{\star}}{\partial t} \left((g - \gamma \P^{\pi'}g)(s,a)\right) \nabla_{\pi'} \int g(s', a') \P(s' \mid s, a) \pi'(a' \mid s')) \big] \\
    & = -\gamma \E_{(s, a) \sim \mu^{\pi_i}_\rho}\big[ \frac{\partial \phi^{\star}}{\partial t} \left((g - \gamma \P^{\pi'}g)(s,a)\right) \E_{\substack{s' \sim \P(\cdot \mid s, a)\\ a' \sim \pi'(\cdot \mid s')}} \left[ g(s', a') \nabla_{\pi'} \log\pi'(a' \mid s')\right]\big] 
\end{align*}

\section{Comparison with AlgaeDICE}
Both the recent AlgaeDICE~\citep{nachum2019algaedice} and our present work propose regularisation based on discounted state-action visitation distribution but in different ways. Firstly, AlgaeDICE is initially designed to find an optimal policy given a batch of training data. They alter the objective function itself i.e the policy performance $J(\pi)$ by adding the divergence between the discounted state-action visitation distribution and training distribution, while our approach adds the divergence term to $L_{\pi}(\pi’)$. The latter is a first order Taylor approximation of the policy performance $J(\pi’)$. Therefore, our approach could be seen as a mirror descent that uses the divergence as a proximity term. Secondly, their training objective is completely different from ours. Their method ends up being an off-policy version of the actor-critic method.

We implemented the AlgaeDICE min-max objective to replace our surrogate min-max objective in the PPO training procedure i.e at each iteration, we sample rollouts from the current policy and update the actor and the critic of AlgaeDICE for 10 epochs. Empirically, we observed that AlgaeDICE objective is very slow to train in this setting. This was expected as it is agnostic to training data while our method leverages the fact that the data is produced by the current policy and estimates advantage using on-policy multi-step Monte Carlo. So our approach is more suitable than AlgaeDICE in this setting. However, AlgaeDICE, as an off-policy method, would be better when storing all history of transitions and updating both actor and critic after each transition, as shown in~\citet{nachum2019algaedice}.

% \newpage
\section{Empirical Results}
\subsection{OpenAI Gym: MuJoCo}
\begin{figure}[h!]
    \centering
    \includegraphics[width=0.3\linewidth]{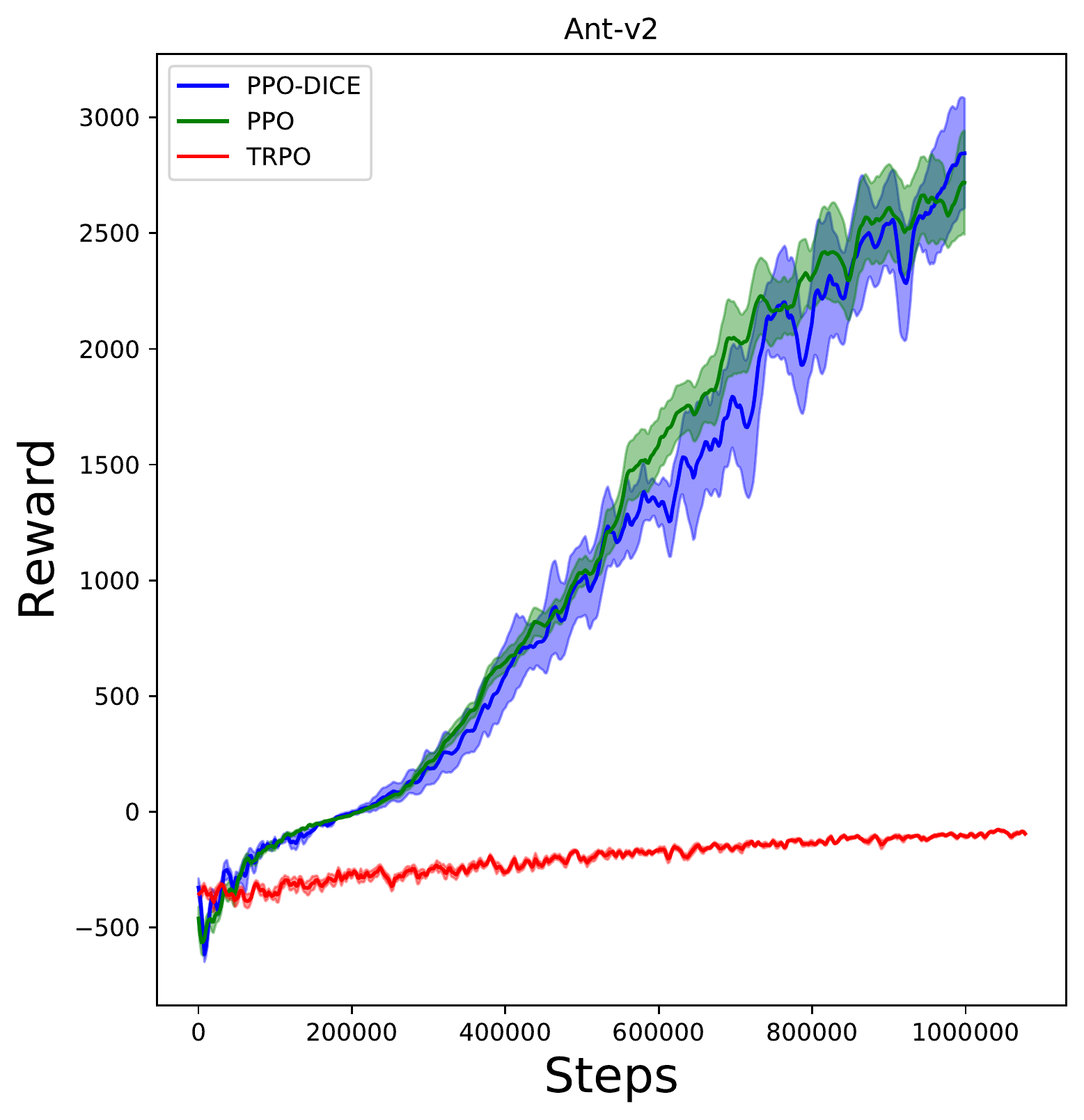}
    \includegraphics[width=0.3\linewidth]{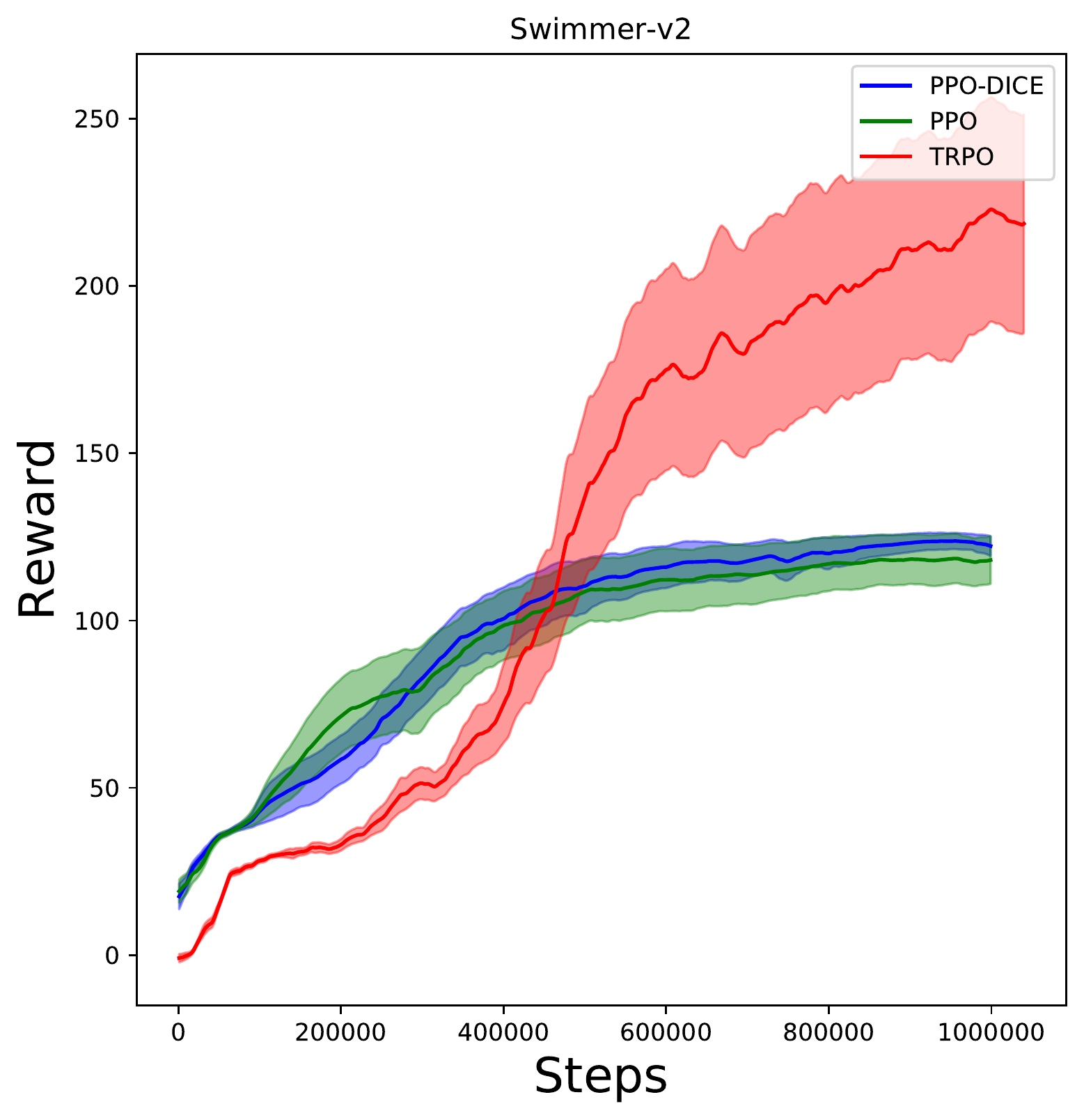}
    \includegraphics[width=0.3\linewidth]{figures/Hopper-v2.pdf}
    \includegraphics[width=0.3\linewidth]{figures/HalfCheetah-v2.pdf}
    \includegraphics[width=0.3\linewidth]{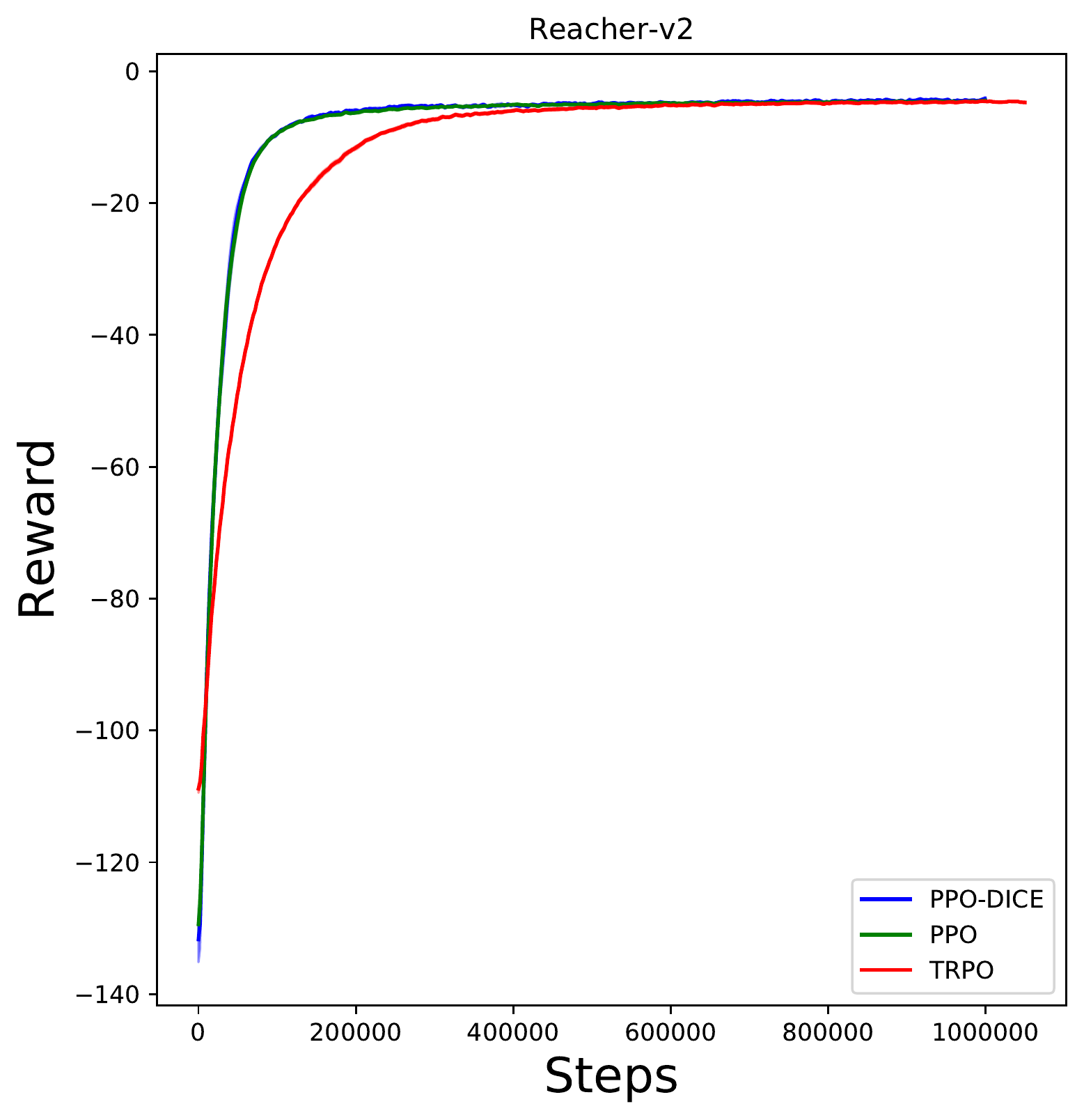}
    \includegraphics[width=0.3\linewidth]{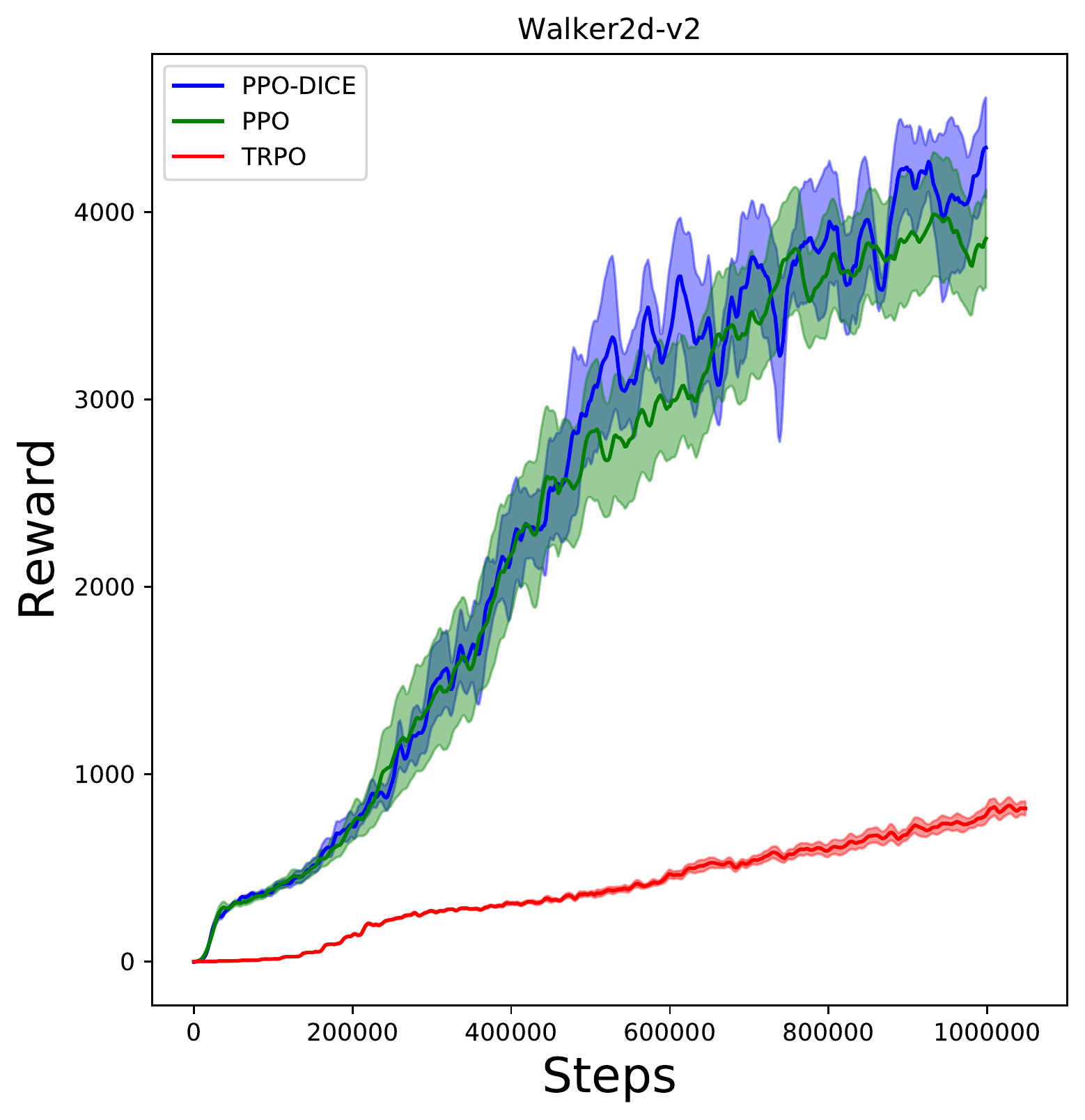}
    \includegraphics[width=0.3\linewidth]{figures/HumanoidStandup-v2.pdf}
    \includegraphics[width=0.3\linewidth]{figures/Humanoid-v2.pdf}
    \includegraphics[width=0.3\linewidth]{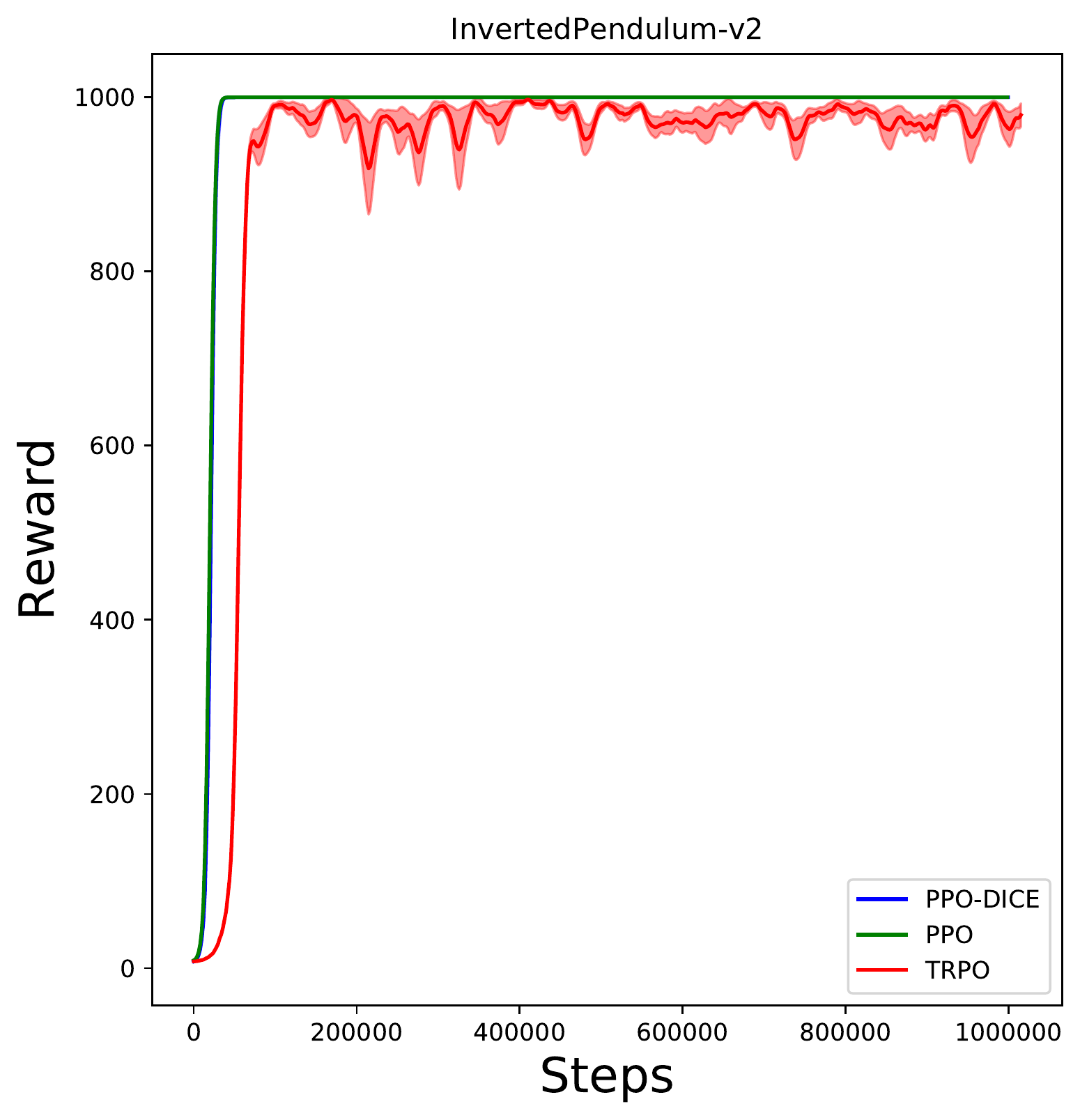}
    \includegraphics[width=0.3\linewidth]{figures/InvertedDoublePendulum-v2.pdf}
    \caption{Our method with KL divergences in comparison to PPO and TRPO on MuJoCo, with 10 seeds. Standard error shaded.}
    \label{fig:dmcontrol}
\end{figure}

See Figure~\ref{fig:dmcontrol}
\label{app:mujoco_results}

\subsection{Atari}
\label{app:atari_results}
\begin{figure}[h!]
    \centering
    \includegraphics[width=0.23\linewidth]{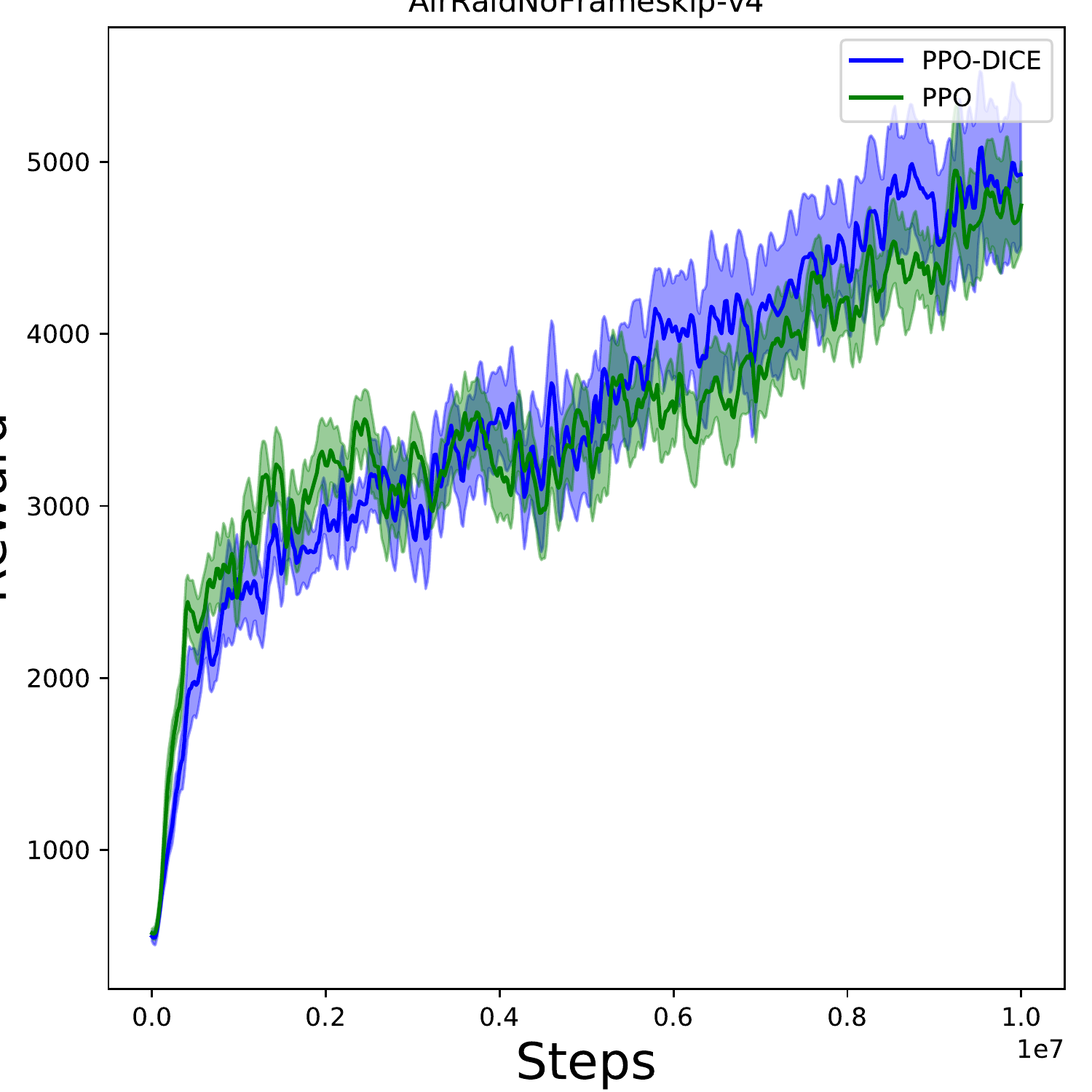}
    \includegraphics[width=0.23\linewidth]{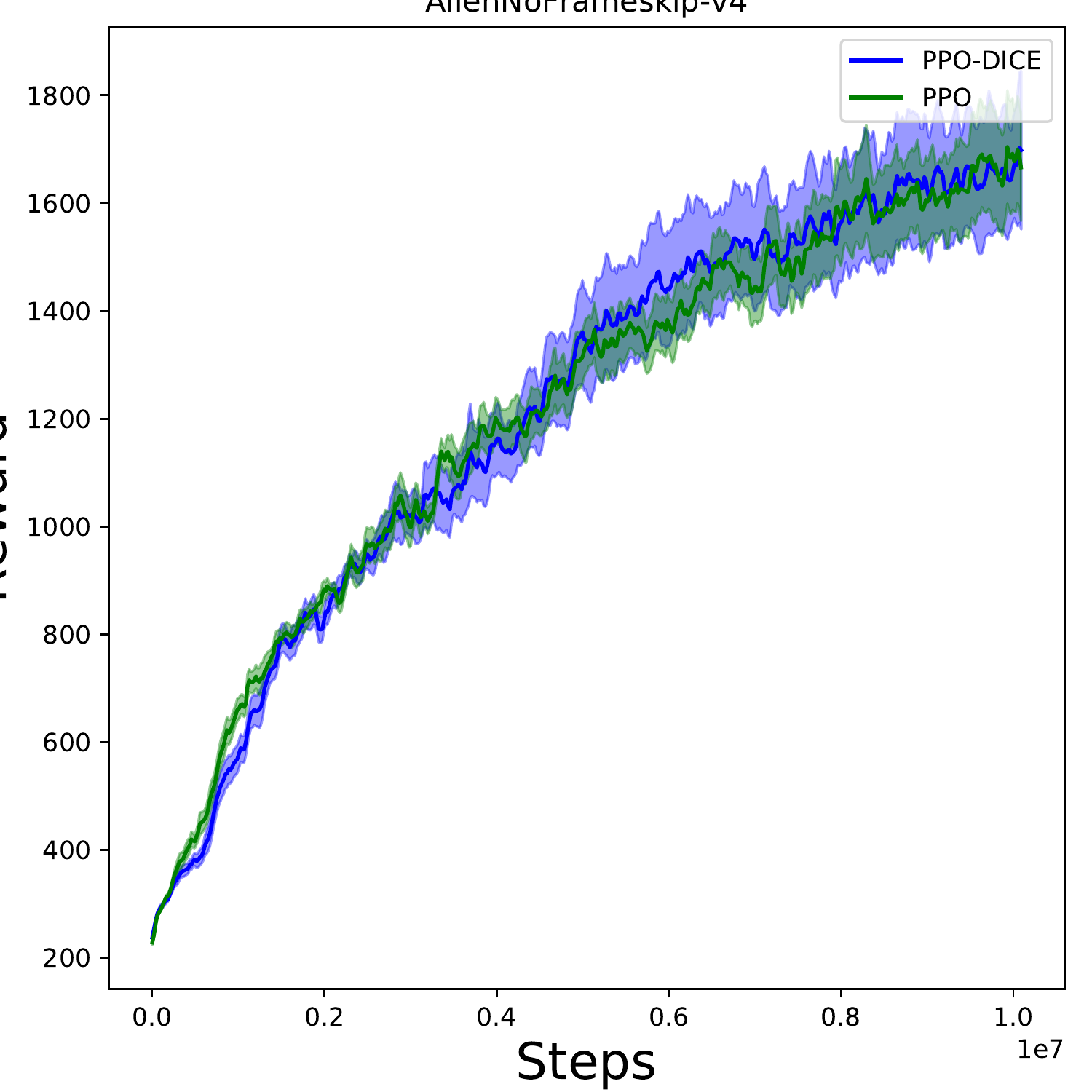}
    \includegraphics[width=0.23\linewidth]{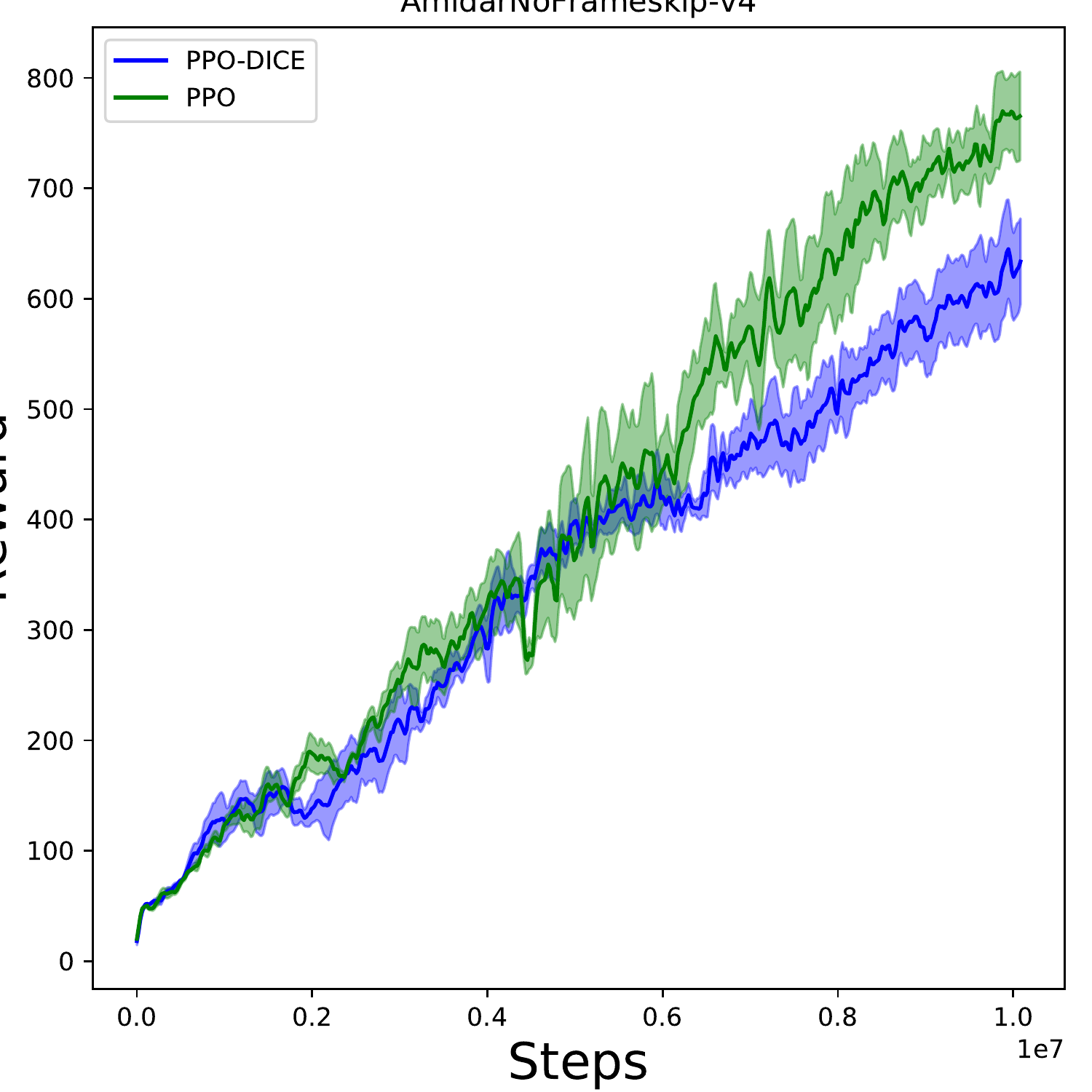}
    \includegraphics[width=0.23\linewidth]{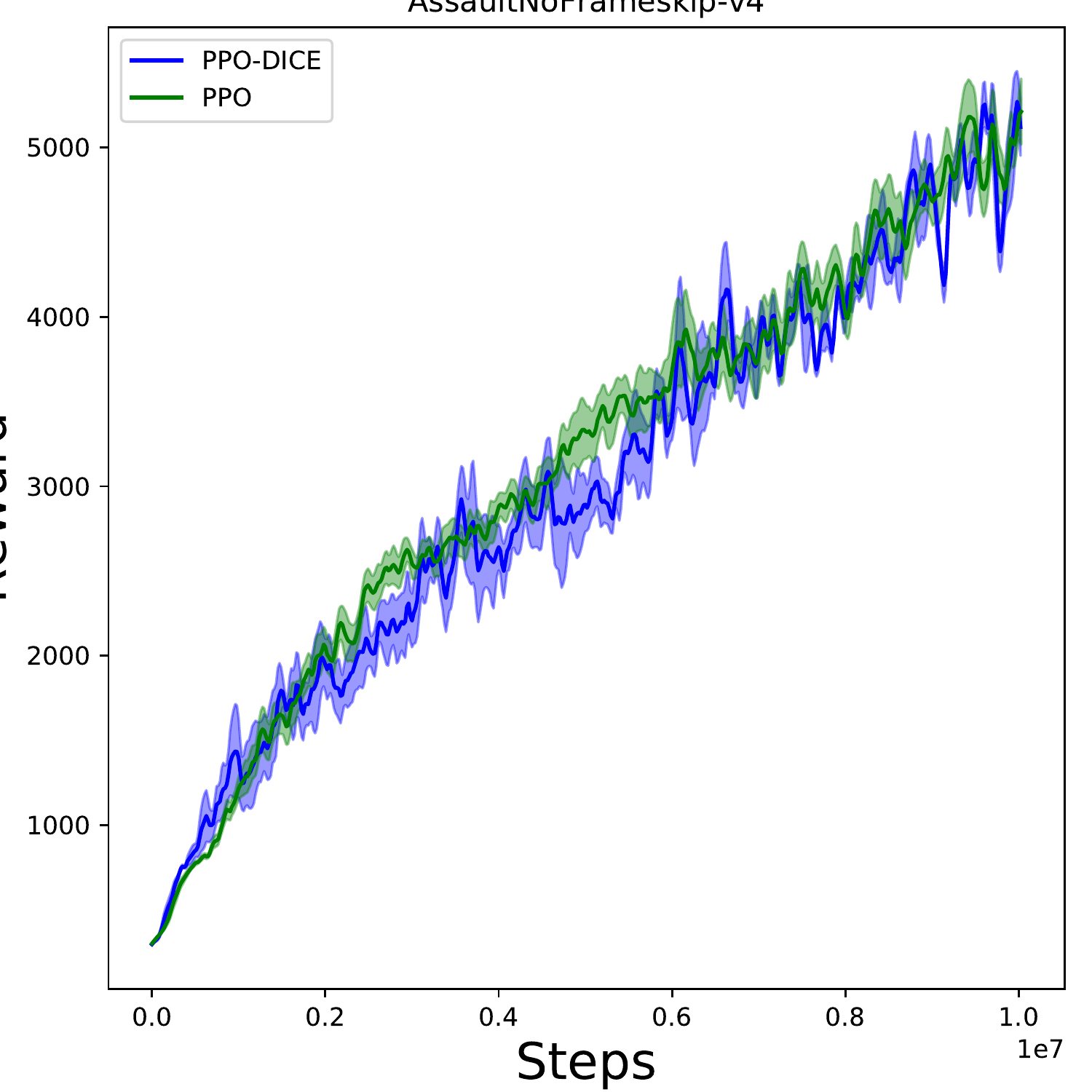}
    \includegraphics[width=0.23\linewidth]{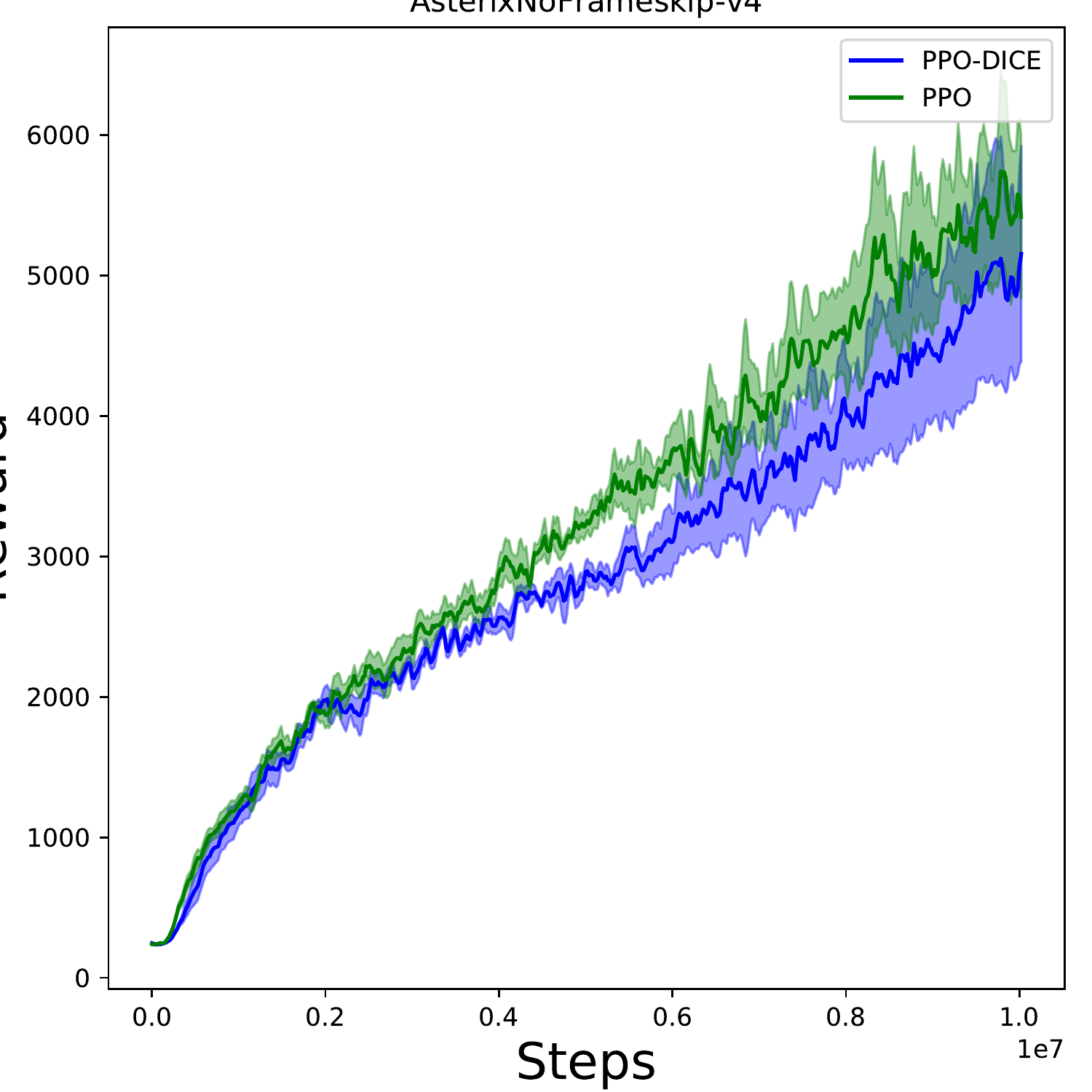}
    \includegraphics[width=0.23\linewidth]{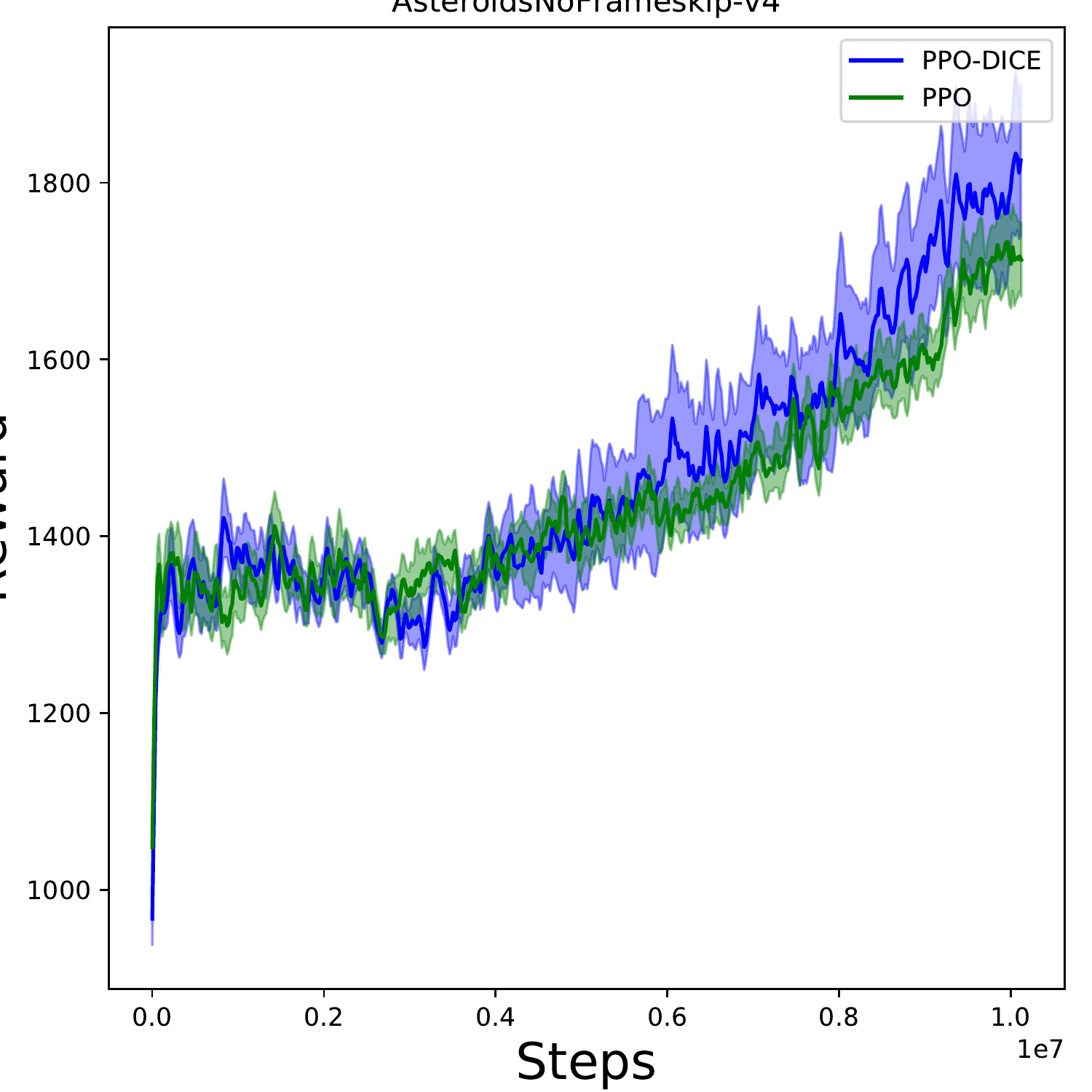}
    \includegraphics[width=0.23\linewidth]{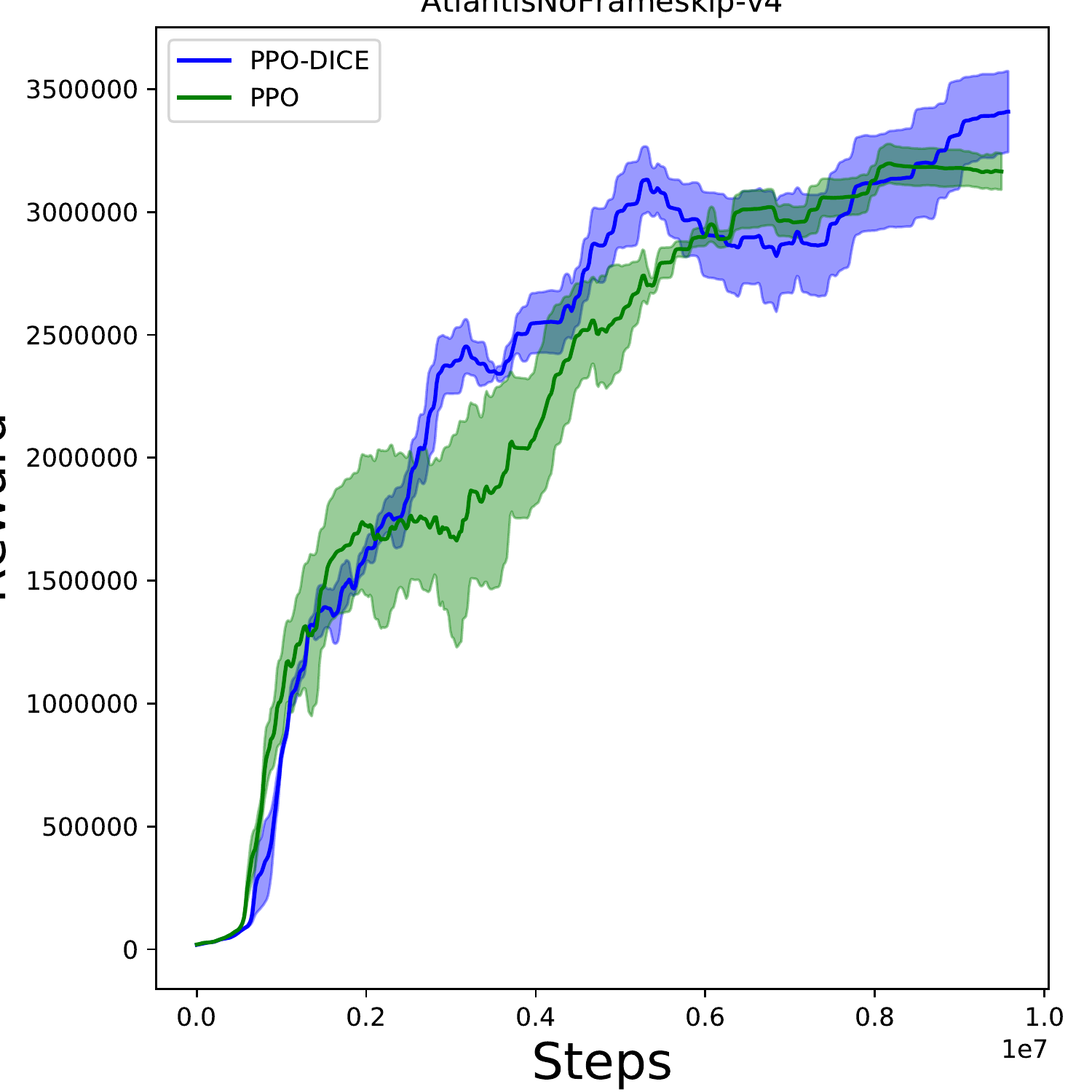}
    \includegraphics[width=0.23\linewidth]{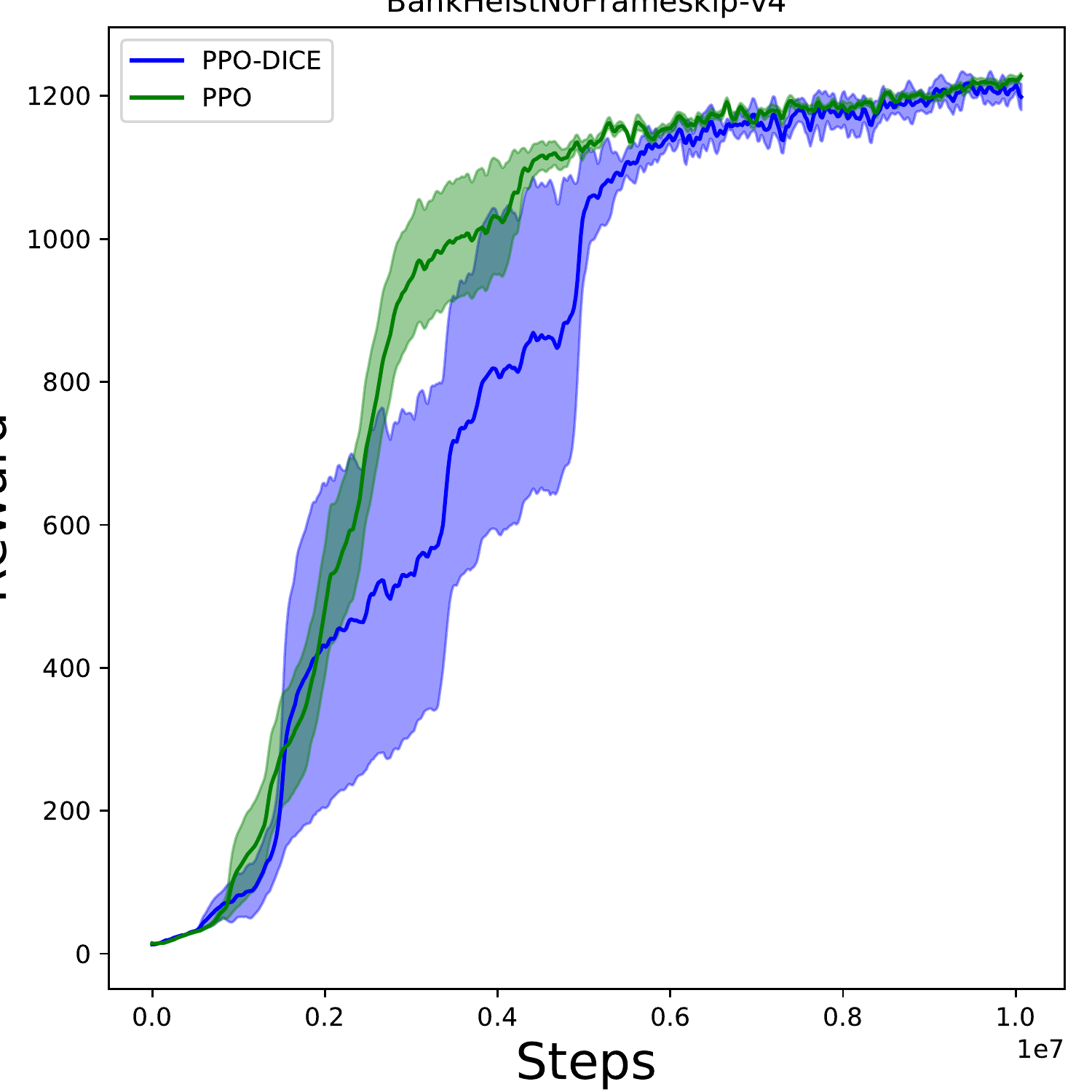}
    \includegraphics[width=0.23\linewidth]{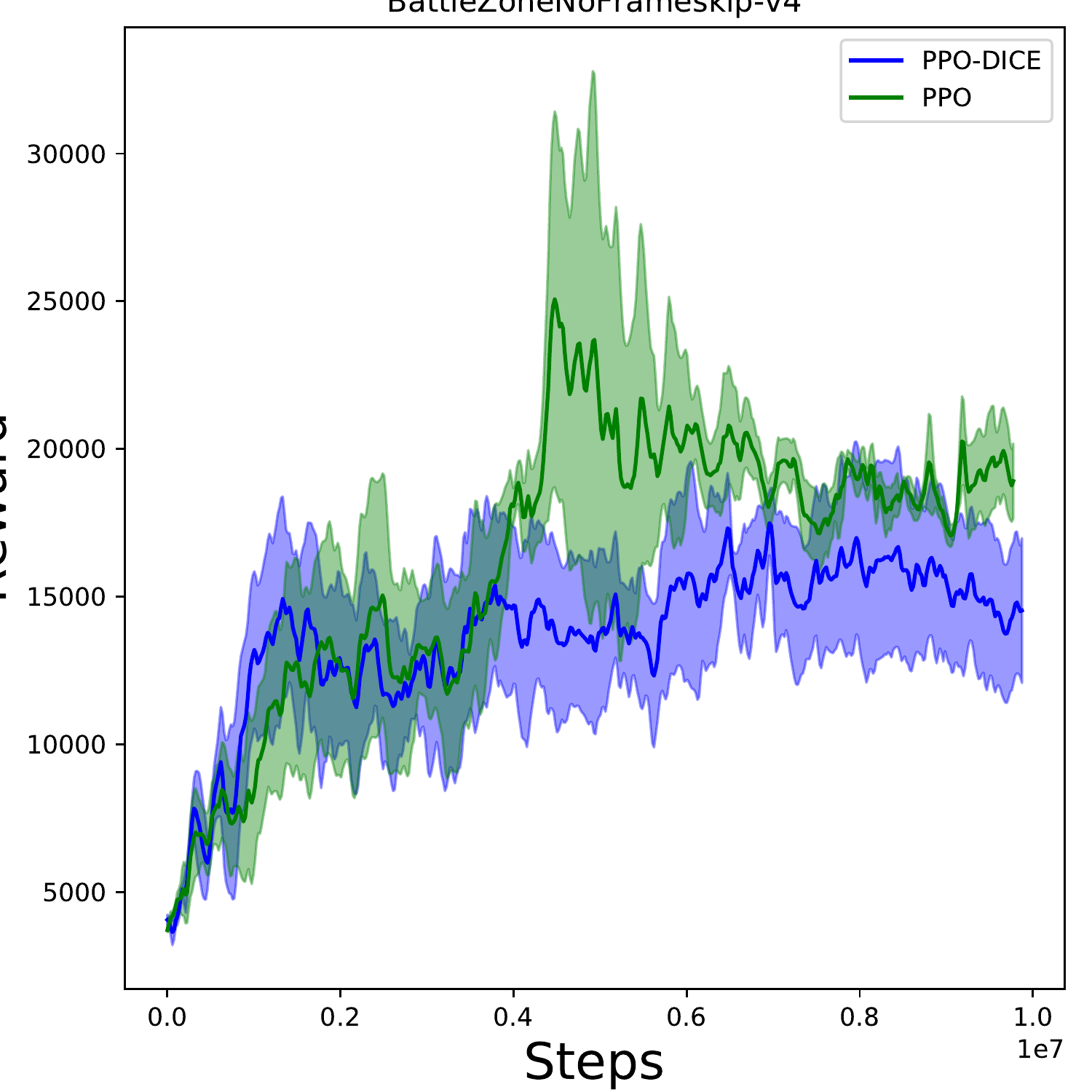}
    \includegraphics[width=0.23\linewidth]{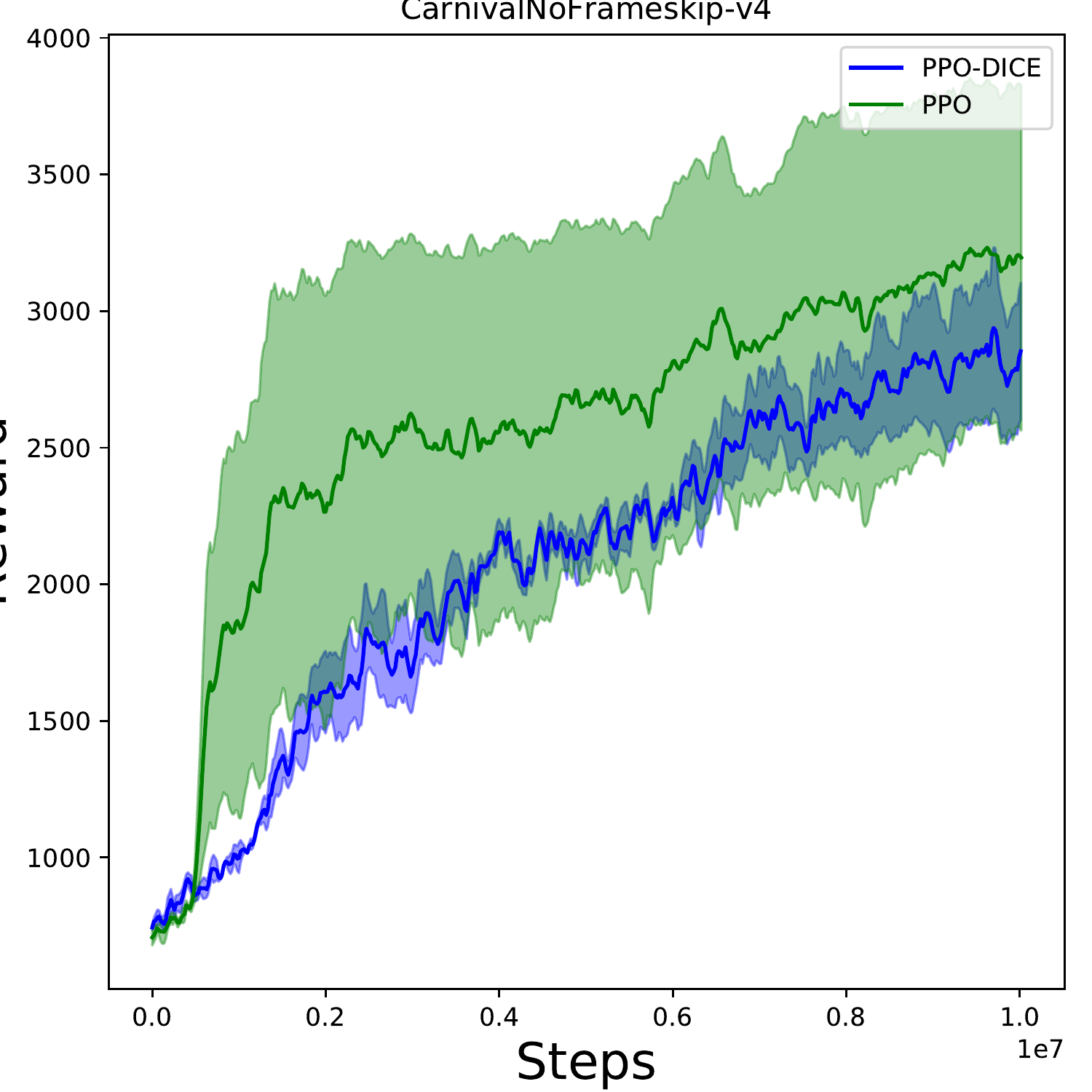}
    \includegraphics[width=0.23\linewidth]{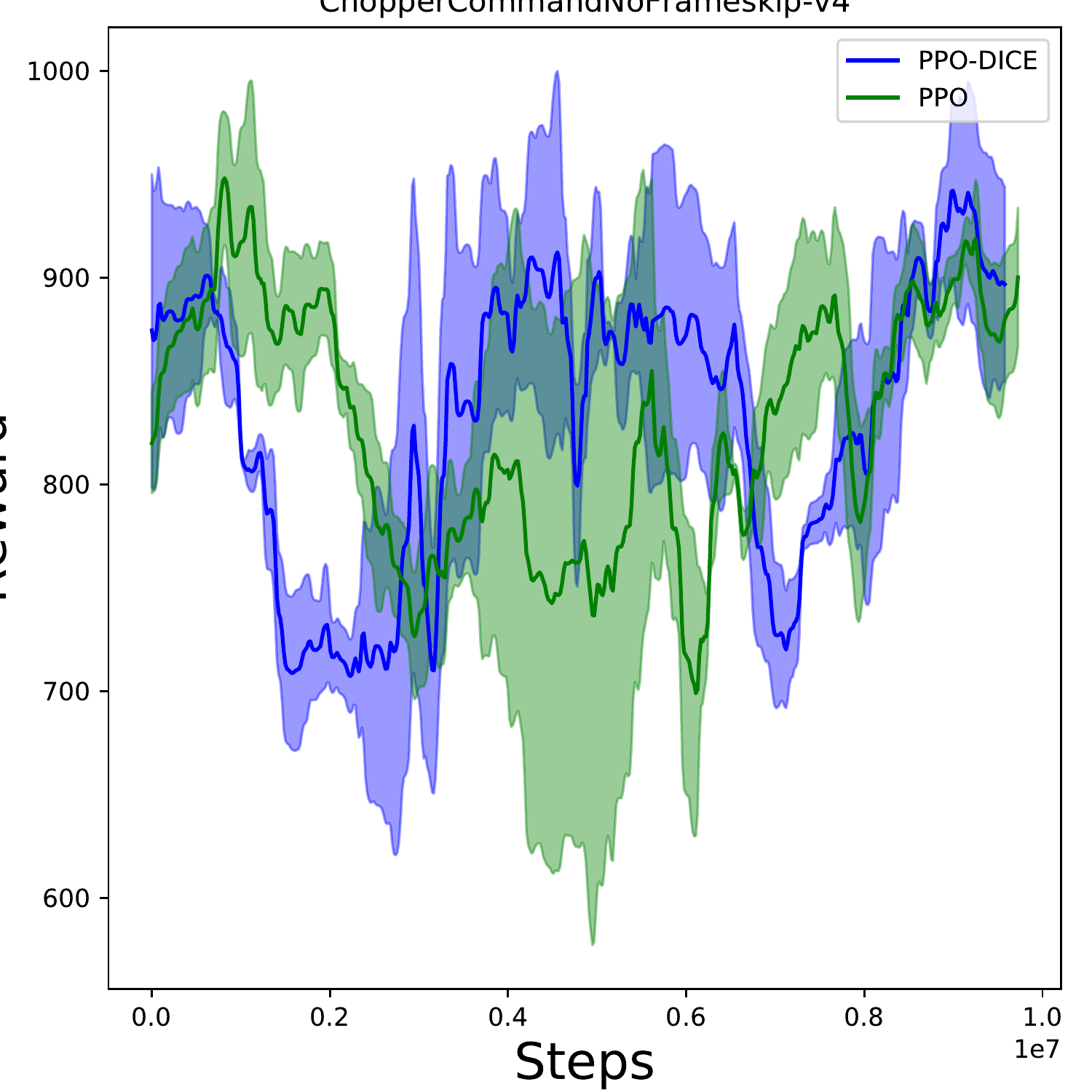}
    \includegraphics[width=0.23\linewidth]{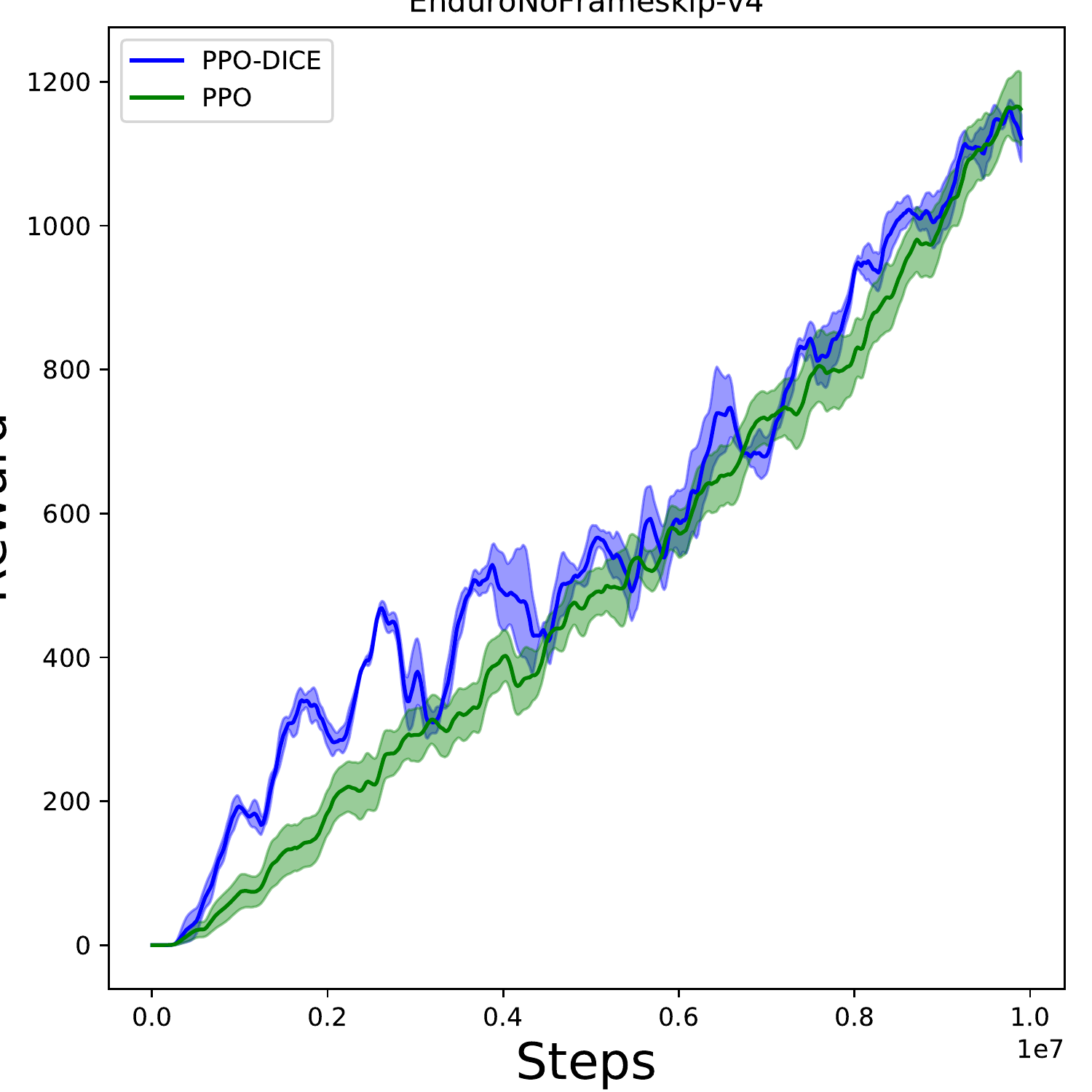}
    \includegraphics[width=0.23\linewidth]{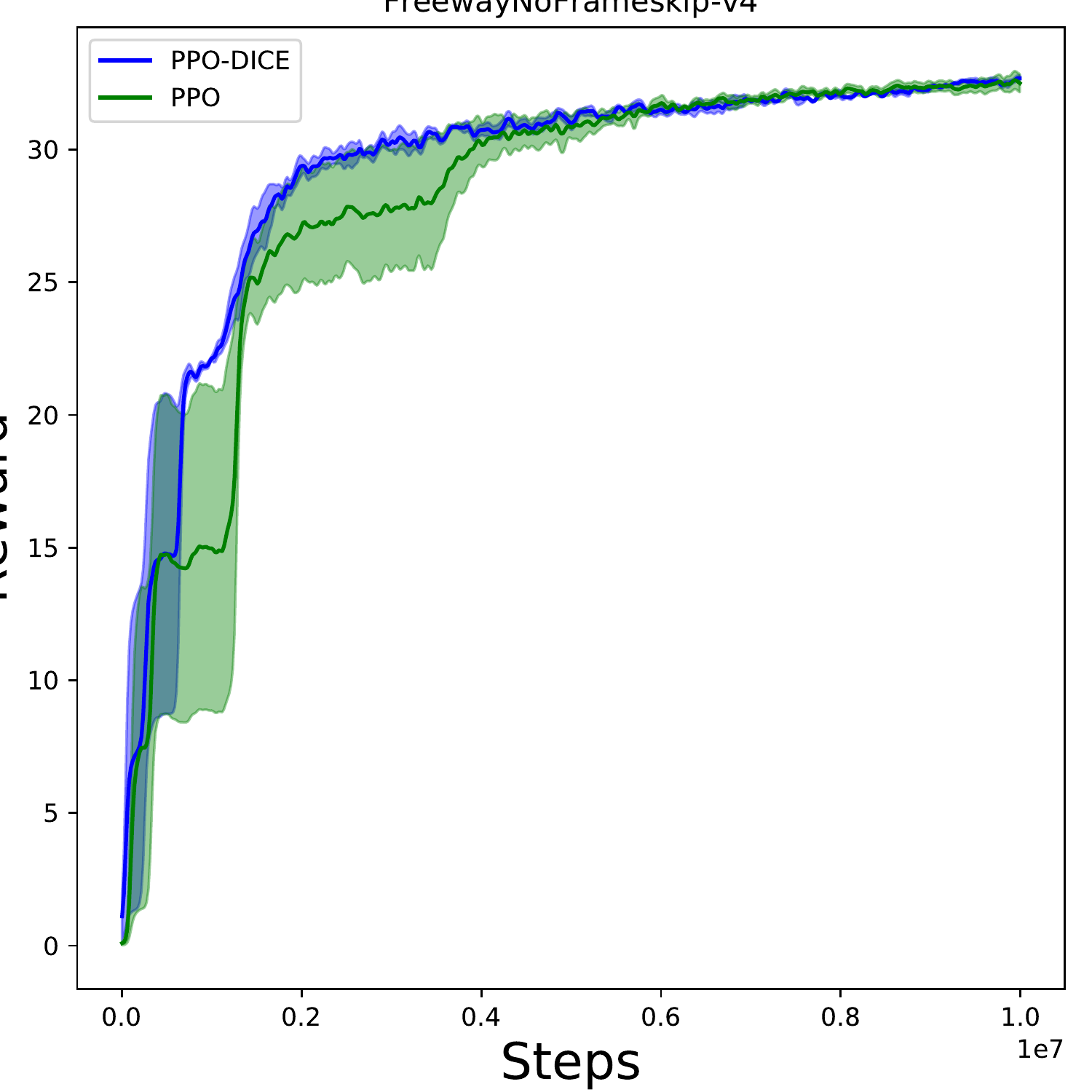}
    \includegraphics[width=0.23\linewidth]{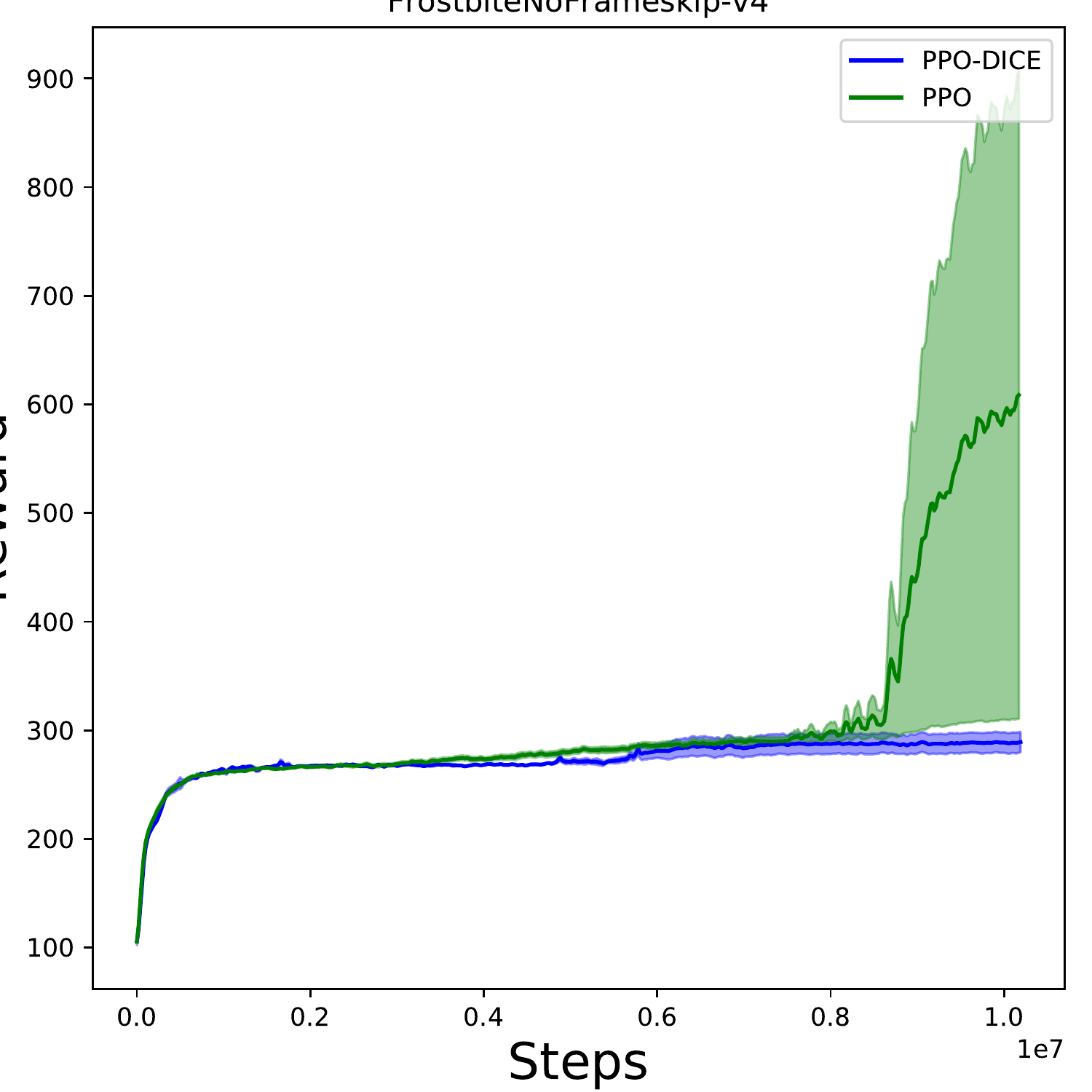}
    \includegraphics[width=0.23\linewidth]{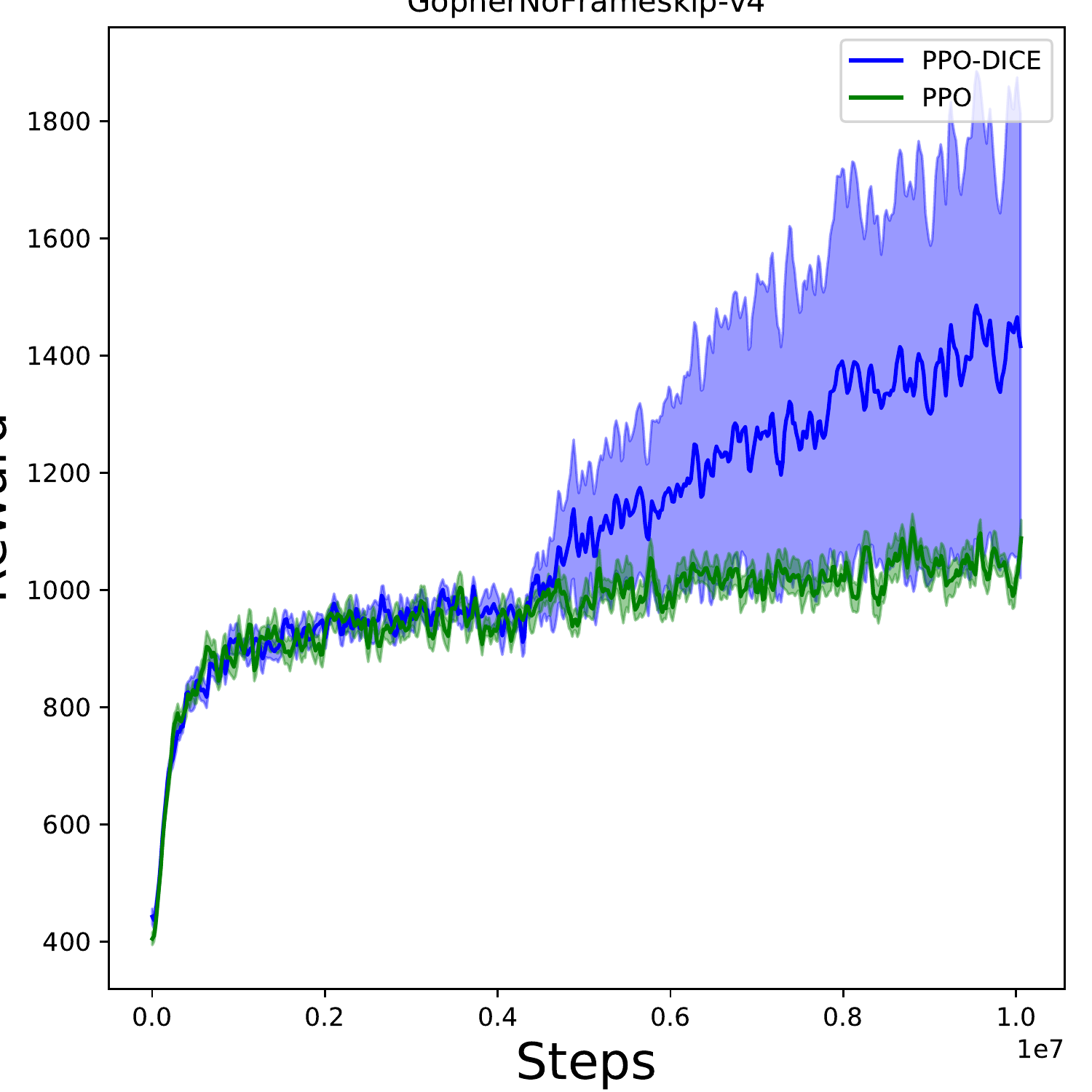}
    \includegraphics[width=0.23\linewidth]{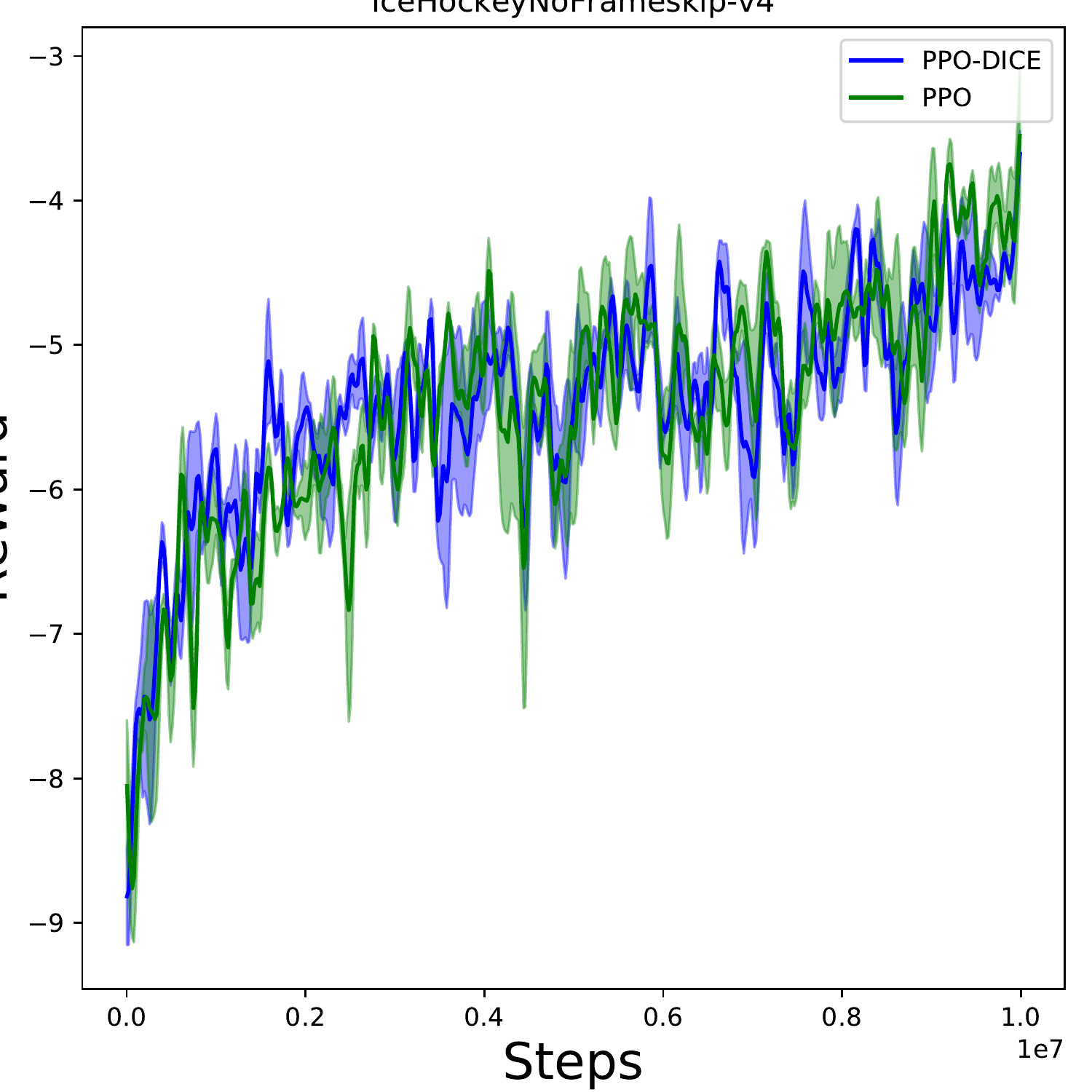}
    \includegraphics[width=0.23\linewidth]{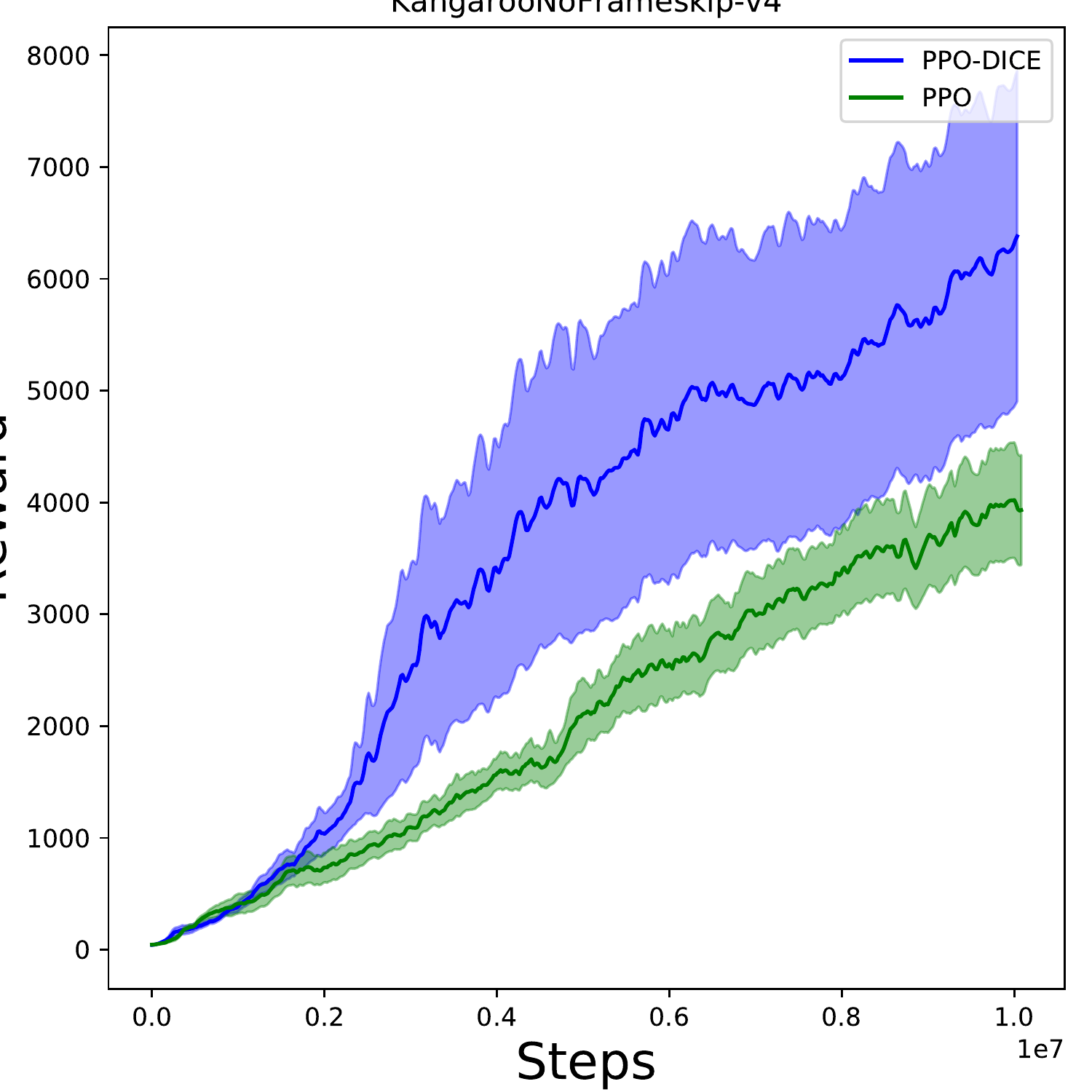}
    \includegraphics[width=0.23\linewidth]{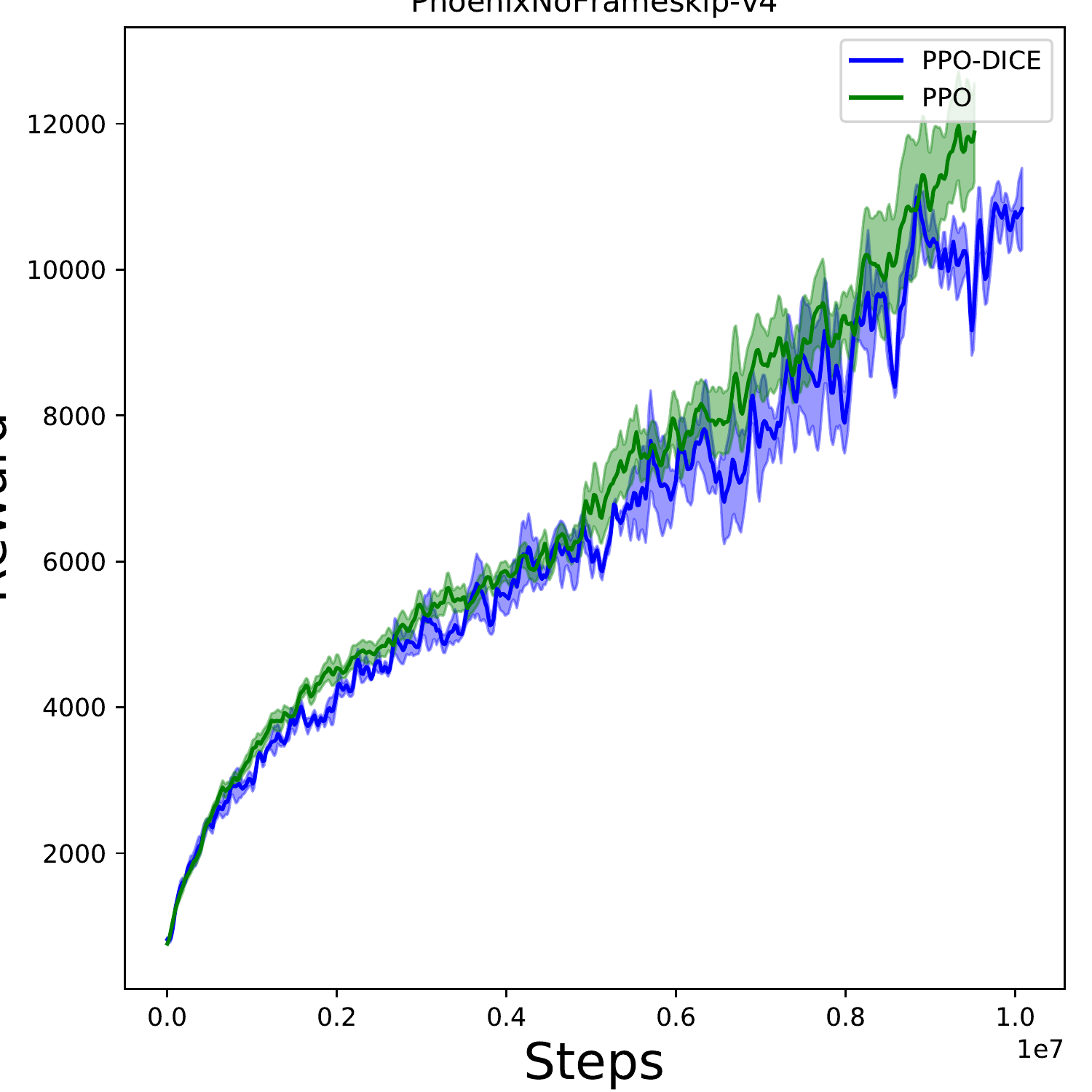}
    \includegraphics[width=0.23\linewidth]{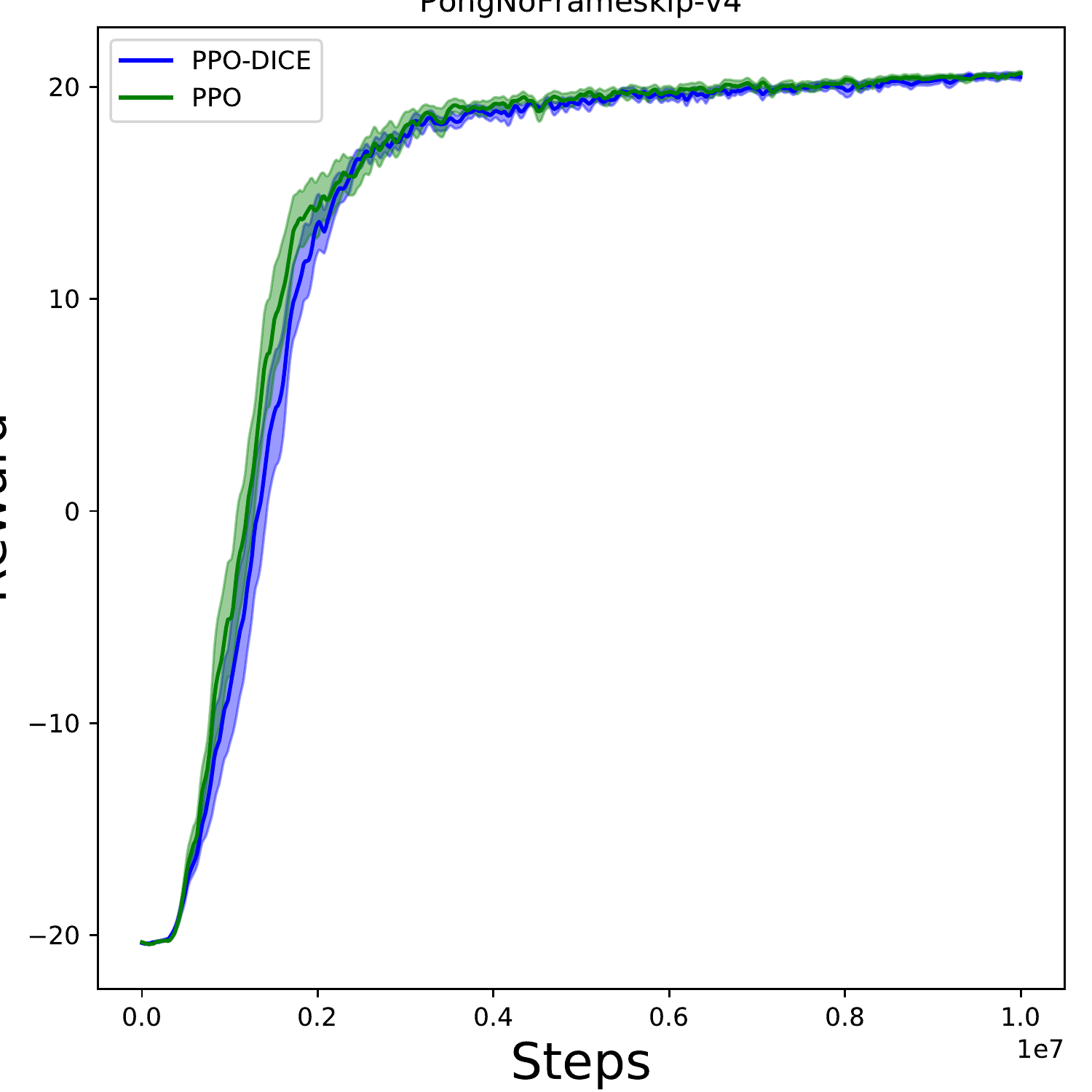}
    \includegraphics[width=0.23\linewidth]{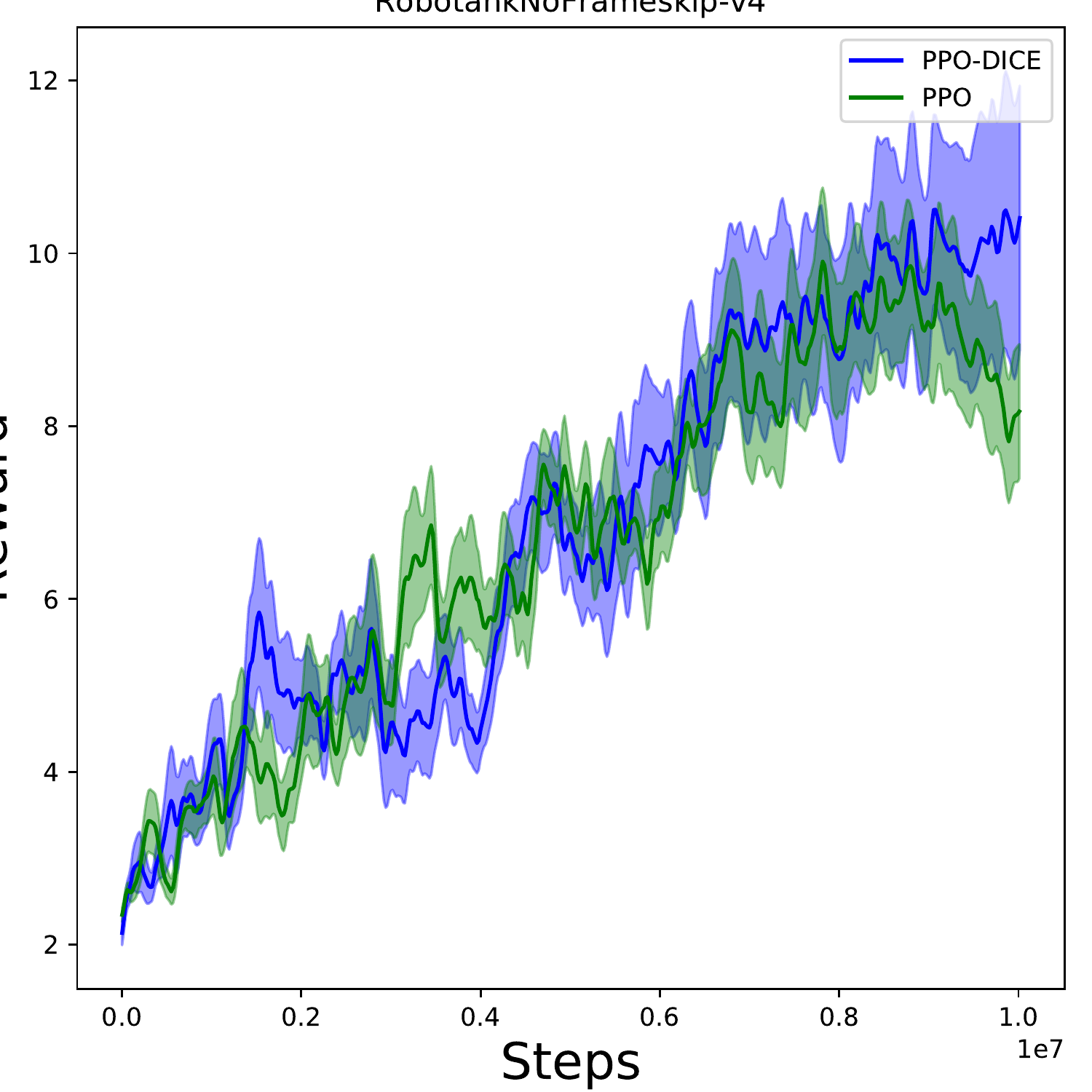}
    \includegraphics[width=0.23\linewidth]{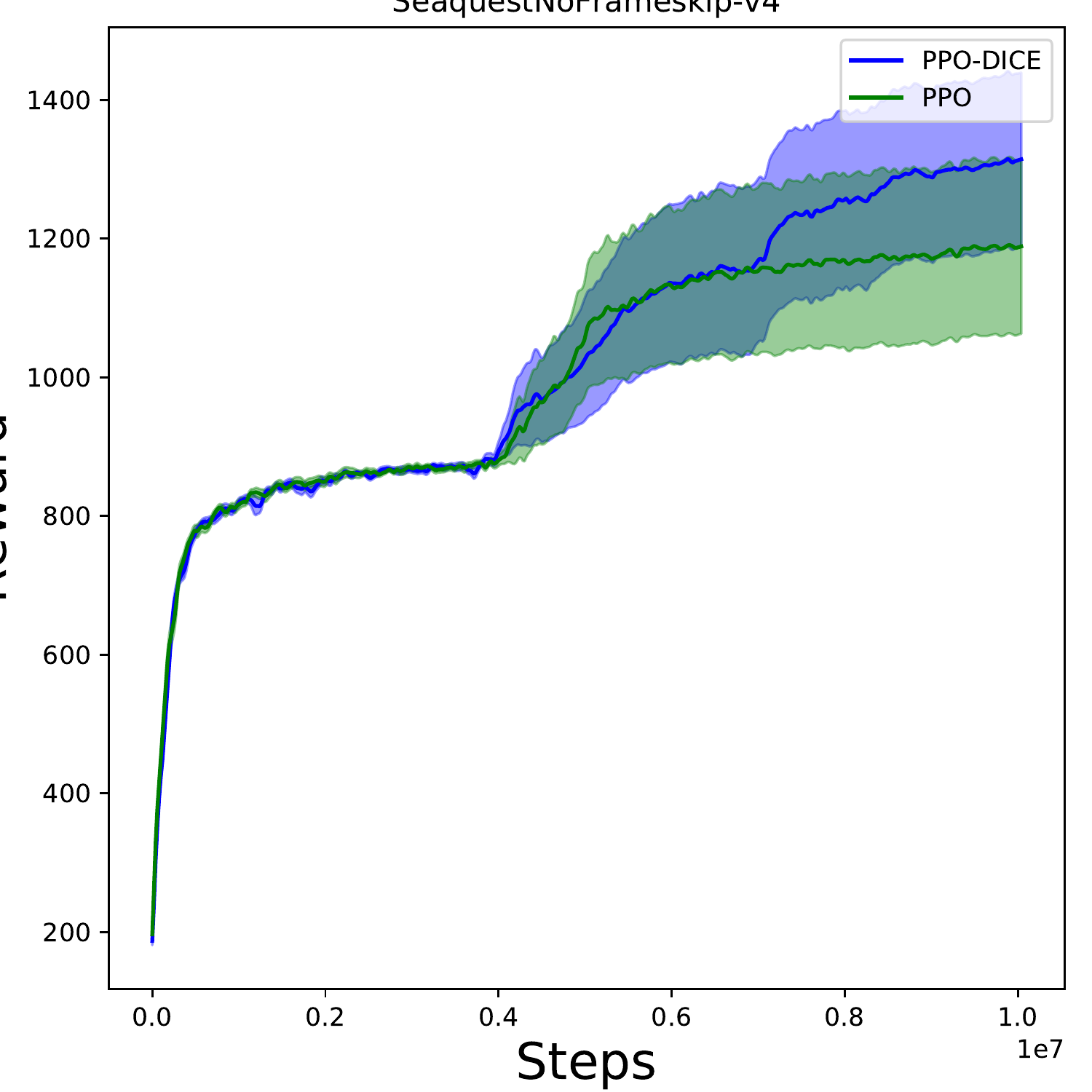}
    \includegraphics[width=0.23\linewidth]{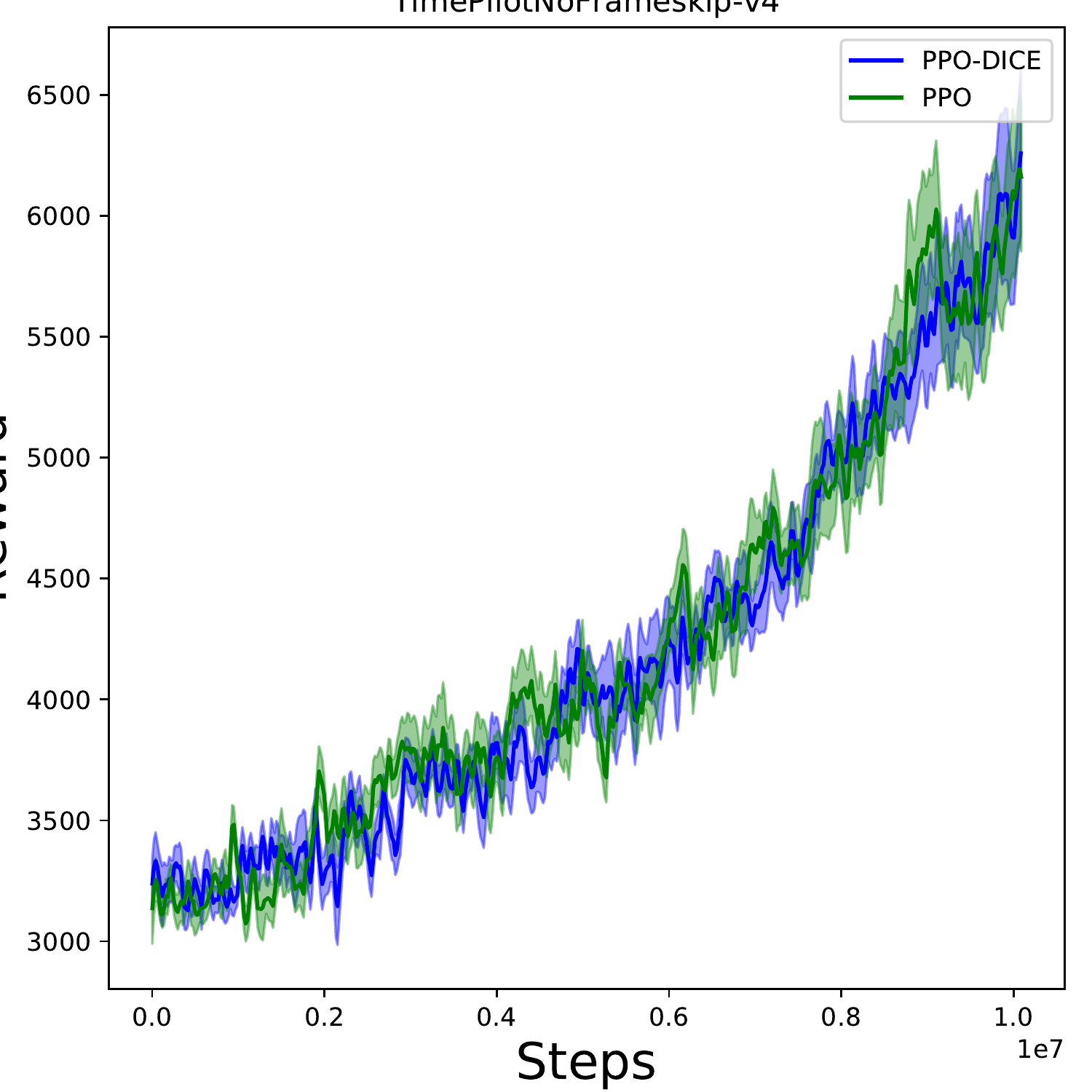}
    \includegraphics[width=0.23\linewidth]{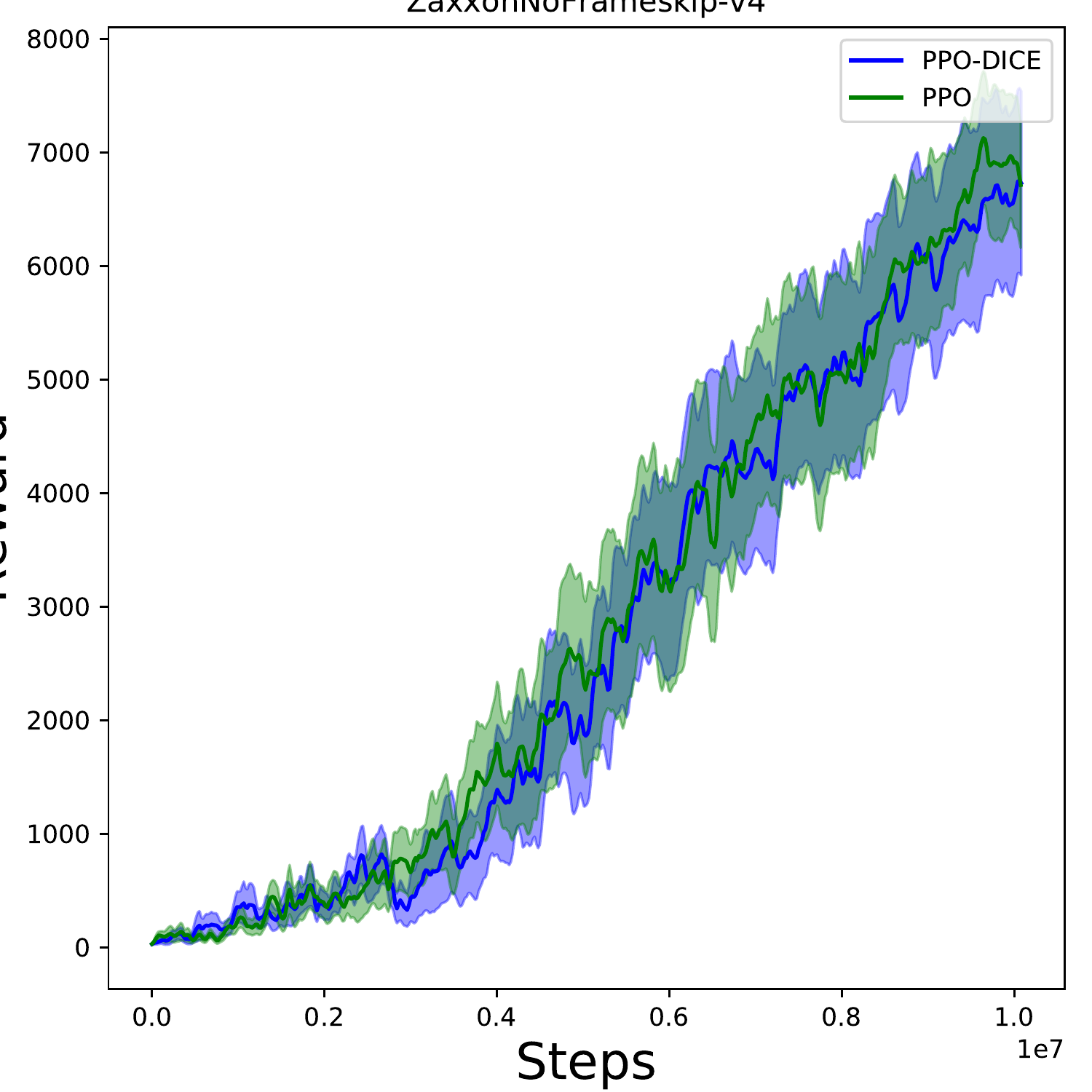}
    \caption{Our method with KL divergences in comparison to PPO on Atari, with 10 seeds and standard error shaded.}
    \label{fig:atari}
\end{figure}

See Figure~\ref{fig:atari}

\section{Hyperparameters}
\label{app:hyperparams}
\subsection{OpenAI Gym: MuJoCo}
For the OpenAI Gym environments we use the default hyperparameters in \citet{pytorchrl}.

\begin{table}[h!]
\centering
\begin{tabular}{|l|c|}
\hline
Parameter name        & Value \\
\hline
Number of minibatches & $4$ \\
Discount $\gamma$ & $0.99$ \\
Optimizer & Adam \\
Learning rate & 3e-4 \\
PPO clip parameter & $0.2$ \\
PPO epochs & 10 \\
GAE $\lambda$ & $0.95$ \\
Entropy coef & $0$ \\
Value loss coef & $0.5$ \\ 
Number of forward steps per update & $2048$ \\
\hline
\end{tabular}\\
\caption{\label{table:hyper_params} A complete overview of used hyper parameters for all methods.}
\end{table}

% For PPO-DICE we used orthonormal regularization on the weights and environment specific parameters for the number of iterations the discriminator was trained per PPO update and the learning rate of the discriminator, which is a multiplicative factor of the PPO learning rate.
% \begin{table}[hb!]
% \centering
% \begin{tabular}{|l|c|c|}
% \hline
% Environment & Disc. Iterations        & Disc. lr factor \\
% \hline
% Ant-v2 & 5 & 2 \\
% HalfCheetah-v2 & 20 & 10 \\
% Hopper-v2 & 5 & 5 \\
% Humanoid-v2 & 5 & 10 \\
% HumanoidStandup-v2 & 5 & 10 \\
% InvertedDoublePendulum-v2 & 5 & 10 \\
% InvertedPendulum-v2 & 5 & 10 \\
% Reacher-v2 &  5 & 10 \\
% Swimmer-v2 & 5 & 5 \\
% Walker2d-v2 & 10 & 1 \\
% \hline
% \end{tabular}\\
% \caption{\label{table:hyper_params} A complete overview of used hyper parameters for PPO-DICE.}
% \end{table}

\subsection{Atari}
For the Atari hyperparameters, we again use the defaults set by \cite{pytorchrl}.

\begin{table}[h!]
\centering
\begin{tabular}{|l|c|}
\hline
Parameter name        & Value \\
\hline
Number of minibatches & $4$ \\
Discount $\gamma$ & $0.99$ \\
Optimizer & Adam \\
Learning rate & 2.5e-4 \\
PPO clip parameter & $0.1$ \\
PPO epochs & 4 \\
Number of processes & $8$ \\
GAE $\lambda$ & $0.95$ \\
Entropy coef & $0.01$ \\
Value loss coef & $0.5$ \\ 
Number of forward steps per update & $128$ \\
\hline
\end{tabular}\\
\caption{\label{table:hyper_params} A complete overview of used hyper parameters for all methods.}
\end{table}

\end{document}